\documentclass[10pt,twocolumn,letterpaper]{article}

\usepackage[algorithms]{wacv}

\usepackage{algorithm}
\usepackage{algpseudocode}

\definecolor{wacvblue}{rgb}{0.21,0.49,0.74}
\usepackage[pagebackref,breaklinks,colorlinks,allcolors=wacvblue]{hyperref}

\def\wacvPaperID{1419}
\def\confName{WACV}
\def\confYear{2027}

\title{Training-Free Debiasing of Diffusion Models via CLIP-Guided Denoising Optimization}

\author{
Dain Kim$^{1,*}$ \quad Jinseo Kim$^{1,*}$ \quad Sungyong Baik$^{1,2 \dagger}$\\
$^{1}$Dept. of Artificial Intelligence, Hanyang University, South Korea\\
$^{2}$Dept. of Data Science, Hanyang University, South Korea\\
{\tt\small \{dain36940, jinseo84, dsybaik\}@hanyang.ac.kr}
}

\begin{document}
\twocolumn[{%
    \maketitle
    \begin{center}
    \includegraphics[width=\linewidth]{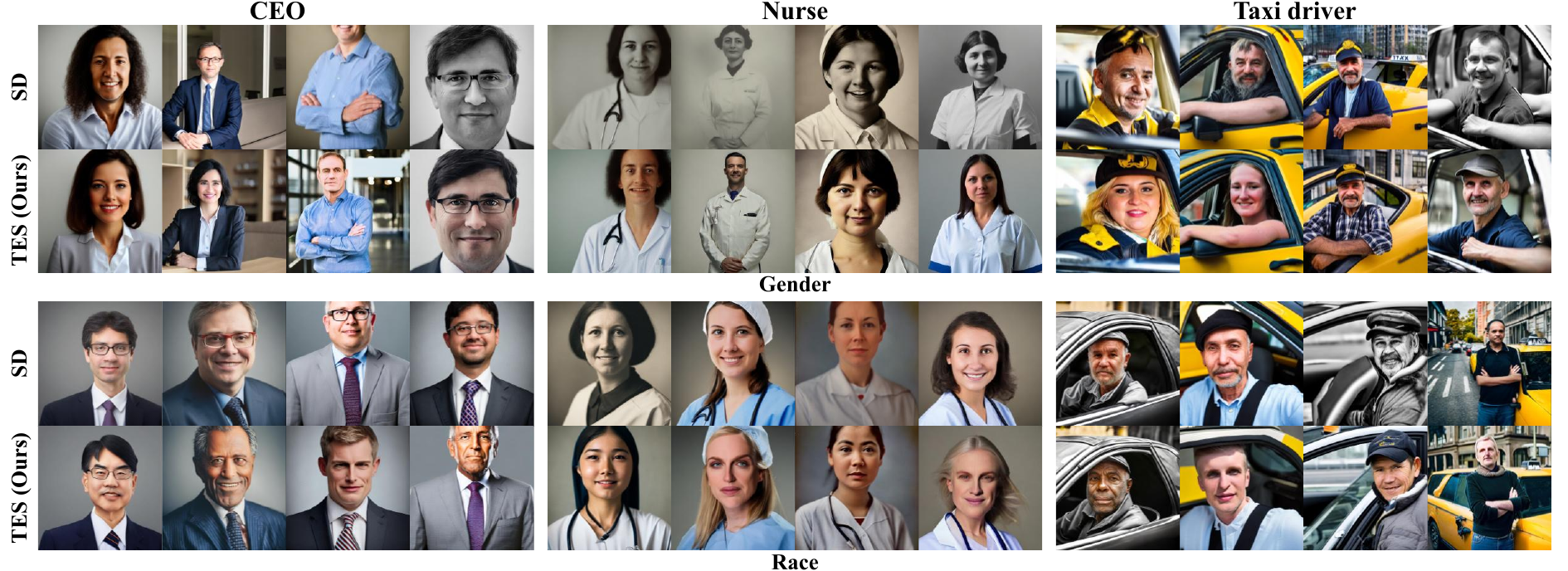}
        \captionof{figure}{\textbf{Overview of qualitative results on Stable Diffusion v2.1.} Our method reduces demographic bias across multiple profession prompts while preserving profession semantics and image quality.
        }
        \label{fig:qualitative}
    \end{center}
}]

\def\thefootnote{*}\footnotetext{Equal contribution.}
\def\thefootnote{${\dagger}$}\footnotetext{Corresponding author.}
\def\thefootnote{\arabic{footnote}}

\begin{abstract}
Text-to-image diffusion models achieve impressive visual quality, yet demographic bias remains a challenge, as neutral prompts consistently produce stereotypical representations across gender and race.
Existing approaches remain limited by costly retraining or by inference-time interventions that often degrade image quality and semantic alignment.
We propose Text Embedding Steering (TES), a training-free framework that mitigates demographic bias by directly optimizing conditional text embeddings during the diffusion process.
We show that a two-stage strategy — early-stage global alignment followed by iterative denoising-time refinement with CLIP-based feedback — enables stable and controllable attribute steering without modifying model parameters. 
Extensive experiments on Stable Diffusion demonstrate that TES outperforms existing training-free baselines in fairness while maintaining competitive image quality.
These results highlight that inference-time text embedding optimization is a practical and scalable solution for fairness-aware generation in diffusion models.
\end{abstract}    
\section{Introduction}
\label{sec:intro}
Recent advances in text-to-image (T2I) models~\cite{sd, pmlr-v139-ramesh21a, NEURIPS2022_ec795aea, nichol2021glide} have enabled the generation of highly realistic and diverse images from natural language prompts.
Models such as Stable Diffusion~\cite{sd} have demonstrated strong performance across a wide range of applications, including content creation, design assistance, and visual storytelling.
As these systems become increasingly integrated into real-world applications, ensuring the reliability and trustworthiness of generated content has emerged as an important challenge.
One critical issue is demographic bias.
Despite receiving neutral prompts, T2I models often generate stereotypical representations associated with attributes such as gender, race, and age.
Such biases can reinforce existing social stereotypes and lead to systematic underrepresentation of certain demographic groups. 
In real-world applications such as advertising, automated content creation, and educational materials, biased outputs can reinforce stereotypes and influence how users perceive social roles.
Moreover, when these models are integrated into downstream pipelines such as data augmentation or content generation systems, bias can accumulate and propagate, reducing the reliability of AI systems.

To address this issue, prior work~\cite{shen2024finetuning, parihar2024balancingact, li2024fairmapping, park2025efa, vardhana2026fully} has explored model retraining, fine-tuning, or inference-time interventions.
While retraining-based approaches can improve fairness, they are computationally expensive and difficult to apply to already deployed models, limiting their practicality.
Inference-time methods provide a more practical alternative by mitigating bias during generation without modifying model parameters. 
However, many existing approaches rely on heuristic prompt engineering or latent-space manipulation, which can be difficult to interpret and control. 
More recent methods attempt to intervene in internal representations through embedding-level adjustments or learned transformations~\cite{moesd, saedebias, li2024fairmapping, fu2025fairimagen}.
While these approaches show promising results, they often depend on predefined directions or static mappings, limiting their ability to provide dynamic and context-aware adjustments during generation.

\begin{figure}[t]
    \centering
    \includegraphics[width=\linewidth]{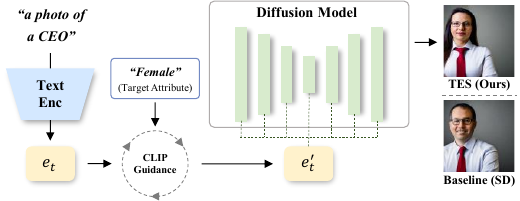}
    \caption{\textbf{Overview of the proposed Text Embedding Steering (TES).} Instead of keeping the text embedding fixed throughout sampling, TES continuously updates the embedding during diffusion, enabling controllable attribute steering while preserving the original prompt semantics.
    }
    \label{fig:1_overview}
\end{figure}

In this work, we propose \textbf{Text Embedding Steering (TES)}, a training-free framework that optimizes text embeddings during the diffusion process.
Our key observation is that demographic attributes are largely determined in the early stages of denoising, where the global structure of the image is formed.
Motivated by this observation, TES performs an early-stage embedding update to guide the overall generation direction, followed by iterative refinement using CLIP-based feedback from intermediate clean-image estimates.
We evaluate TES on a diverse set of profession-based prompts and show that it effectively reduces gender and race bias while preserving image quality and semantic alignment.
Importantly, improvements in metrics correspond to more balanced and realistic representations of social roles, leading to more reliable and trustworthy outputs in real-world applications. An overview of our approach is illustrated in Figure~\ref{fig:1_overview}.
Given an input prompt, our method dynamically updates the text embedding during the diffusion process, enabling continuous and context-aware control of generated attributes. 

Our contributions are summarized as follows:
\begin{itemize}
    \item We propose TES, a training-free framework that mitigates demographic bias by continuously optimizing text embeddings without requiring model retraining.
    \item We identify the importance of early-stage intervention and introduce a two-stage optimization strategy for stable and consistent control.
    \item We demonstrate that TES achieves a strong balance between fairness and image quality, and improves the reliability of generated content in practical deployment scenarios.
\end{itemize}
\section{Related Work}
\subsection{Bias in Text-to-Image Diffusion Models}
Recent studies have shown that text-to-image (T2I) diffusion models~\cite{Cho_2023_ICCV, wang-etal-2023-t2iat, NEURIPS2023_b01153e7, Bianchi_2023, wu2024stablediffusionexposedgender, culturalbias} inherit and often amplify social biases, resulting in demographically skewed or stereotypical visual representations. 
In particular, profession-related prompts exhibit strong demographic associations, where occupations such as \textit{CEO} or \textit{engineer} are predominantly associated with male representations
~\cite{NEURIPS2023_b01153e7, Bianchi_2023, wu2024stablediffusionexposedgender, perera2023analyzing, jain2020imperfect}.
Prior work further shows that these biases affect not only demographic frequencies but also generation quality and semantic fidelity across groups~\cite{wang-etal-2023-t2iat, seshadri2023biasamplificationparadoxtexttoimage}.
These findings suggest that demographic bias in T2I models is systematic rather than incidental, motivating the need for controllable and reliable debiasing methods.

\subsection{Training-Based Debiasing Methods}
Training-based approaches mitigate bias through data rebalancing, fine-tuning, or fairness-aware optimization~\cite{Liu_2024_CVPR, li2024fairmapping, parihar2024balancingact, shrestha2024fairrag, hou2026aitti,lightfair,vanbreugel2021decaf,teo2023fairtl}. Representative methods include SCoFT~\cite{Liu_2024_CVPR}, Fair Mapping~\cite{li2024fairmapping}, Balancing Act~\cite{parihar2024balancingact}, FairRAG~\cite{shrestha2024fairrag}, AITTI~\cite{hou2026aitti}, and LightFair~\cite{lightfair}.
While effective, these methods require access to training data or model parameters, often involving costly fine-tuning procedures.
As a result, they are difficult to apply to closed-source or already deployed models, limiting their practical usability and motivating training-free alternatives.

\subsection{Training-Free Debiasing Methods}
To overcome these limitations, recent work~\cite{fairdiffusion, entigen, zhang2023itigen, debiaspi, jiang2024laag, choi2024fairsm, yesiltepe2024mist, fairgen, orgad2023time} has explored training-free debiasing strategies that operate entirely at inference time.
These methods can be broadly categorized into three groups.
Prompt-based approaches modify input prompts to encourage more balanced generations~\cite{fairdiffusion, entigen, debiaspi}.
While simple and easy to deploy, they are heuristic and often sensitive to prompt wording and sampling conditions.
Sampling-based methods instead intervene during the denoising process through mechanisms such as attention guidance, cross-attention editing, switching strategies, or latent guidance~\cite{jiang2024laag, choi2024fairsm, yesiltepe2024mist, fairgen}.
For example, FairGen~\cite{fairgen} discovers demographic control directions through an additional learning stage and applies them during inference.
Although this enables plug-and-play debiasing, it still requires a separate adapter training stage.
Embedding-based methods instead operate directly in the representation space.
FairImagen~\cite{fu2025fairimagen} mitigates demographic bias by projecting prompt embeddings onto a predefined fairness-aware subspace before generation.
However, these methods rely on fixed embedding transformations or predefined control directions that remain unchanged throughout sampling.

Unlike previous approaches, TES formulates debiasing as a dynamic optimization problem over conditional text embeddings.
Rather than relying on separately trained adapters, or predefined demographic control directions, TES continuously updates the conditional text embedding using CLIP-based feedback from intermediate clean-image estimates, enabling adaptive and context-aware demographic control throughout denoising without retraining or auxiliary modules.
\section{Preliminaries}
\subsection{Diffusion Models}
Denoising Diffusion Probabilistic Models (DDPM)~\cite{ddpm} generate images by progressively transforming Gaussian noise into data samples through a sequence of denoising steps.
Starting from a random latent variable $x_T \sim \mathcal{N}(0, I)$, the model iteratively removes noise to obtain a clean sample.
At each timestep $t$, a neural network $\epsilon_\theta$ predicts the noise component in the latent variable, and the sample is updated according to a predefined diffusion schedule:
\begin{equation}
x_{t-1} = f(x_t, \epsilon_\theta(x_t, t)).
\end{equation}
In text-to-image diffusion models, the denoising network is conditioned on text embeddings obtained from a pretrained text encoder.
The conditional noise prediction can be written as $\epsilon_\theta(x_t, t, e)$, where $e$ denotes the text embedding that encodes the input prompt.
Latent diffusion models such as Stable Diffusion~\cite{sd} further improve efficiency by performing diffusion in a compressed latent space.

Among different sampling methods, DDPM follows a stochastic reverse process, while Denoising Diffusion Implicit Models (DDIM)~\cite{ddim} provide a deterministic formulation:
\begin{equation}
x_{t-1} = \sqrt{\bar{\alpha}_{t-1}} \, \hat{x}_0 + \sqrt{1 - \bar{\alpha}_{t-1}} \, \epsilon_\theta(x_t, t).
\end{equation}
Compared to stochastic sampling, DDIM produces more stable intermediate states and more predictable denoising trajectories.
This property is particularly beneficial for optimization-based generation methods, since stable trajectories enable reliable feedback and iterative updates during sampling.
Following prior inference-time optimization approaches, we adopt DDIM sampling throughout this work.

\subsection{Clean Image Estimation}
Given a noisy latent $x_t$ and the predicted noise $\epsilon_\theta(x_t, t)$, the clean sample $\hat{x}_0$ can be estimated as:
\begin{equation}
\label{eq:x0}
\hat{x}_0 = \frac{x_t - \sqrt{1 - \bar{\alpha}_t} \, \epsilon_\theta(x_t, t)}{\sqrt{\bar{\alpha}_t}}.
\end{equation}
This formulation provides an estimate of the final denoised image at each timestep.
Compared to the noisy latent $x_t$, the predicted clean sample $\hat{x}_0$ offers a more interpretable approximation of the underlying image content and often reveals semantic structure before the denoising process is completed.
As a result, clean-image estimates have been widely used in diffusion-based guidance and optimization methods as a proxy for image-level supervision.
In this work, we leverage $\hat{x}_0$ as a semantic feedback signal for updating conditional text embeddings during inference.
By evaluating intermediate clean-image estimates rather than noisy latents, the optimization objective can operate directly on semantically meaningful visual attributes.
\section{Method}
\begin{figure*}[t]
    \centering
    \includegraphics[width=\textwidth]{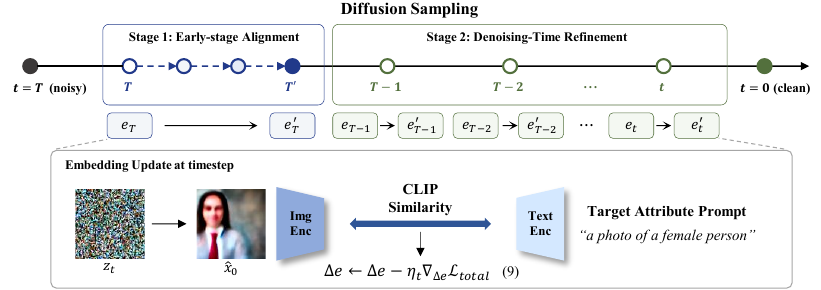}
    \caption{
        \textbf{Overall framework of Text Embedding Steering (TES).} TES first performs an early-stage embedding alignment to steer the global generation trajectory, followed by denoising-time refinement that iteratively updates the text embedding using CLIP feedback from predicted clean images. At each active timestep, the reconstructed clean image $\hat{x}_0$ is compared with the target attribute prompt, and the resulting objective updates the embedding perturbation. This enables continuous and context-aware demographic steering while preserving the original prompt semantics.
    }
    \label{fig:overview}
\end{figure*}

\subsection{Overview}
Given a neutral text prompt $T_{\text{base}}$, such as ``a photo of a CEO'', our goal is to generate images that preserve the semantic content of the prompt while reducing demographic imbalance in the generated outputs. 
Following prior work, we formulate fairness-aware generation as a demographic distribution alignment problem, where the generated demographic distribution is encouraged to match a predefined target distribution. 
In this paper, we mainly adopt a uniform target distribution across demographic groups, although TES can support arbitrary target distributions depending on the application.
Figure~\ref{fig:overview} illustrates the overall TES pipeline with early-stage alignment and denoising-time refinement.

Let $G_\theta$ denote a frozen text-to-image diffusion model and $e_0$ the conditional text embedding produced by the text encoder.
Instead of updating the model parameters, TES optimizes only an additive perturbation to the conditional text embedding,
\begin{equation}
e_t = e_0 + \Delta e,
\end{equation}
while keeping the diffusion backbone, text encoder, and sampling procedure fixed. 
TES dynamically updates this embedding throughout denoising using CLIP-based feedback from intermediate clean-image estimates. 
This enables adaptive demographic control without retraining or auxiliary modules, while preserving the semantic content of the original prompt.

\subsection{CLIP-Guided Embedding Optimization}
At each optimization step, TES reconstructs an intermediate clean image and evaluates it using CLIP~\cite{clip}.
Let $I_t$ denote the decoded clean-image estimate at timestep $t$.
The target attribute used during optimization is determined by the automatic target-assignment procedure, with details provided in the supplementary material.
Consequently, the optimization objective is derived from the predefined target distribution rather than from manually specified demographic labels.

\noindent\textbf{Embedding Objective.}
We define an attribute alignment objective using a target attribute description $T_{\text{target}}$ and an opposing attribute description $T_{\text{opp}}$.
For gender debiasing, these correspond to descriptions such as
``a male person'' and ``a female person.'' Let
$s_{\text{target}}=\mathrm{sim}(I_t,T_{\text{target}})$ and
$s_{\text{opp}}=\mathrm{sim}(I_t,T_{\text{opp}})$
denote the CLIP cosine similarities to the target and opposite attribute prompts, respectively.
The attribute loss is
\begin{equation}
\mathcal{L}_{\text{attr}}
=
-\log
\frac{\exp(s_{\text{target}}/\tau)}
{\exp(s_{\text{target}}/\tau)+\exp(s_{\text{opp}}/\tau)},
\end{equation}
where $\tau$ is the temperature.
To preserve the semantic content of the original prompt, we further optimize
\begin{equation}
    \mathcal{L}_{\text{prof}}=1-\mathrm{sim}(I_t,T_{\text{base}}).
\end{equation}
To prevent excessive deviation from the original embedding, we additionally regularize the embedding perturbation,
\begin{equation}
    \mathcal{L}_{\text{reg}}=\|\Delta e\|_2^2.
\end{equation}
The complete objective is
\begin{equation}
    \mathcal{L}_{\text{total}}=\lambda_{\text{attr}}\mathcal{L}_{\text{attr}}+\lambda_{\text{prof}}\mathcal{L}_{\text{prof}}+\gamma\mathcal{L}_{\text{reg}}.
\end{equation}
This objective encourages the generated image to move toward the predefined demographic distribution while preserving the semantic content of the original prompt and preventing excessive embedding drift.

\noindent\textbf{Intermediate Clean-Image Supervision.}
Rather than computing CLIP supervision directly from noisy latent representations, TES evaluates the reconstructed clean image estimated at each denoising step.
Given the noisy latent $x_t$ and the predicted noise $\epsilon_\theta(x_t,t,e_t)$, we first estimate the clean sample $\hat{x}_0$, decode it into image space, and compute the CLIP-guided objective using the resulting image $I_t$.
Optimization is intentionally deferred from the earliest denoising steps because the reconstructed clean-image estimate remains unreliable under extremely high noise levels.
Instead, embedding updates are activated only within a selected timestep window, where the reconstructed image already captures meaningful semantic structure while the global generation trajectory remains sufficiently flexible for effective intervention.

\subsection{Two-Stage Optimization}
TES performs embedding optimization in two complementary stages: an early-stage alignment followed by denoising-time refinement.

\noindent\textbf{Stage 1: Early-stage Alignment.}
The first stage provides a coarse global correction before demographic attributes become firmly established.
Rather than directly modifying the diffusion model, TES performs gradient updates on the conditional text embedding using the CLIP-guided objective,
\begin{equation}
    \Delta e\leftarrow\Delta e-\eta_t\nabla_{\Delta e}\mathcal{L}_{\mathrm{total}}.
\end{equation}
This initial update biases the global generation trajectory toward the desired demographic distribution before fine-grained visual details emerge.

\noindent\textbf{Stage 2: Denoising-Time Refinement.}
After the initial alignment, TES continuously refines the conditional embedding during subsequent denoising steps.
At each optimization timestep, the embedding is updated using the same CLIP-guided objective computed from the reconstructed clean image.
This iterative refinement adapts the embedding to the evolving image content while maintaining the demographic alignment established during the first stage, preventing the generation from drifting back toward biased outcomes.

\begin{figure}[t]
    \centering
    \includegraphics[width=\linewidth]{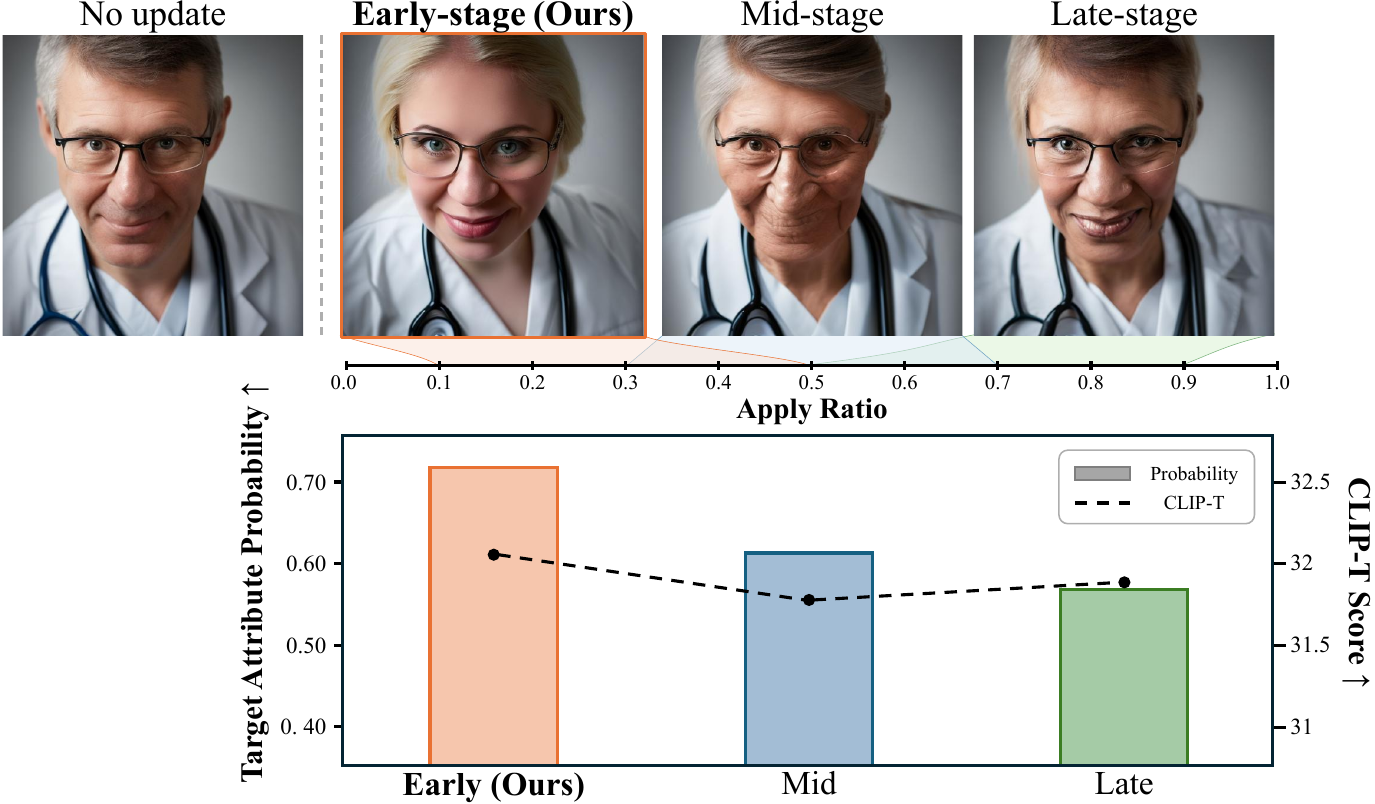}
    \caption{\textbf{Effect of embedding update timing.} 
    Apply Ratio denotes the fraction of total denoising steps during which embedding updates are active, where 0.1–0.5 corresponds to early-stage, 0.3–0.7 to mid-stage, and 0.5–0.9 to late-stage intervention.}
    \label{fig:timing}
\end{figure}

\noindent\textbf{Active Timestep Window.}
Embedding updates are activated only within a predefined timestep window, $t\in[t_{\mathrm{start}},t_{\mathrm{end}})$.
The window is chosen to intervene early enough to influence the global generation trajectory, while avoiding the earliest steps where the reconstructed clean-image estimate is still dominated by noise.
The update-timing analysis in Figure~\ref{fig:timing} supports this design, showing that early intervention enables more coherent semantic shifts, whereas later updates mainly affect localized details after the trajectory has been established.
Thus, the selected window provides a practical trade-off between semantic reliability and controllability.

\noindent\textbf{Learning-Rate Schedule.}
Within the active window, TES adopts a timestep-dependent learning rate,
\begin{equation}
    \eta_t=\eta\cdot\phi(t),
\end{equation}
where $\eta$ denotes the base learning rate and $\phi(t)$ is a schedule over the active window.
Unless otherwise specified, we employ a linearly increasing schedule,
\begin{equation}
    \phi(t)=\frac{t_{\mathrm{start}}-t}{t_{\mathrm{start}}-t_{\mathrm{end}}}.
\end{equation}
This schedule applies smaller updates when the reconstructed image is still relatively uncertain and gradually increases the update strength as more reliable semantic information becomes available.
The influence of the timestep window and learning-rate schedule is analyzed in the ablation study.

\begin{algorithm}[t]
    \caption{Inference-time text embedding optimization}
    \label{alg:method}
    \begin{algorithmic}[1]
        \State \textbf{Input:} base prompt $T_{\text{base}}$, initial embedding $e_0$
        \State \textbf{Initialize:} $\Delta e \gets 0$, $x_T \sim \mathcal{N}(0, I)$
        \For{$t = T, \ldots, 1$}
            \State compute CFG prediction using $e = e_0 + \Delta e$
            \State update latent $x_t \rightarrow x_{t-1}$
            \If{$t \in [t_{\mathrm{start}}, t_{\mathrm{end}})$}
                \State estimate clean sample $\hat{x}_0$ using Eq.~\eqref{eq:x0}
                \State decode $\hat{x}_0$ into image $I$
                \For{$k = 1, \ldots, K$}
                    \State compute $\mathcal{L}_{\mathrm{attr}}$ and $\mathcal{L}_{\mathrm{prof}}$
                    \State form total loss $\mathcal{L}_{\mathrm{total}}$
                    \State update $\Delta e \gets \Delta e - \eta_t \nabla_{\Delta e}\mathcal{L}_{\mathrm{total}}$
                \EndFor
            \EndIf
        \EndFor
        \State \textbf{Return:} generated image
    \end{algorithmic}
\end{algorithm}

\subsection{Integration with Diffusion Sampling}
TES is integrated into the standard classifier-free guidance pipeline without modifying the diffusion backbone, text encoder, or sampler.
At each denoising timestep, the current conditional embedding, $e_t=e_0+\Delta e$, is first used to compute the standard classifier-free guidance prediction, and the latent is updated accordingly.
If the current timestep lies outside the active optimization window, sampling proceeds normally.
Otherwise, TES reconstructs the intermediate clean image from the updated latent, evaluates the CLIP-guided objective, and updates the embedding perturbation before proceeding to the next denoising step.
Consequently, the embedding used at timestep $t-1$ incorporates the feedback obtained from the reconstructed image at timestep $t$.
Algorithm~\ref{alg:method} summarizes the complete inference-time procedure.
Unlike one-shot embedding editing, TES continuously refines the conditional text embedding throughout sampling while leaving all model parameters unchanged.
As a result, the proposed framework remains fully training-free and can be directly applied to existing diffusion models without additional retraining or auxiliary modules.

\section{Experiments}
\begin{table*}[t]
    \caption{\textbf{Quantitative results on gender attributes using Stable Diffusion v2.1.}~\cite{sd}
    Best and second-best results are indicated by \textbf{bold} and \underline{underline}.}
    \label{tab:main}
    \resizebox{\linewidth}{!}{
    \begin{tabular}{c|ccc|ccccc}
        \toprule
        & \multicolumn{8}{c}{\textbf{Gender}}\\
        \cmidrule(lr){2-9} 
        
        \textbf{Method}
        & \multicolumn{3}{c|}{\textbf{Fairness}}
        & \multicolumn{5}{c}{\textbf{Quality}} \\
        
        & Bias-O$\downarrow$ & Bias-Q$\downarrow$
        & FD$\downarrow$ & CLIP-T$\uparrow$ & CLIP-I$\uparrow$ & DINO$\uparrow$ & FID$\downarrow$ & IS$\uparrow$\\
        \midrule
        
        SD\cite{sd}
        & 0.85 (±0.05) & 1.84 (±0.63) & 0.54 (±0.03) & 29.90 (±0.15) & - & - & 259.36 (±4.81) & 1.23 (±0.03)\\
        
        debias VL\cite{chuang2023debiasvl}
        & 0.43 (±0.09) & 1.44 (±0.48) &  0.28 (±0.05) & 28.20 (±0.22) & 70.01 (±0.96) & 0.49 (±0.02) & 245.11 (±3.72) & 1.35 (±0.03)\\
        
        UCE\cite{gandikota2024uce}
        & 0.90 (±0.04) & 1.67 (±0.71) &  0.59 (±0.02) & 29.41 (±0.13) & \textbf{87.94} (±0.86) & \textbf{0.71} (±0.02) & 268.52 (±3.92) & 1.22 (±0.02)\\
        
        EntiGen\cite{entigen}
        & 0.42 (±0.03) & 2.10 (±0.38) &  0.25 (±0.02) & 29.25 (±0.16) & 69.22 (±1.12) & 0.49 (±0.02) & 255.01 (±3.60) & 1.24 (±0.02)\\
        
        LightFair\cite{lightfair}
        & \underline{0.33} (±0.10) & \underline{1.40} (±0.28) & 
         \underline{0.21} (±0.05) & \textbf{30.82} (±0.19) & 75.29 (±0.99) & 0.63 (±0.02) & \underline{231.46} (±3.30) & \underline{1.35} (±0.02)\\
        
        \midrule
        \textbf{TES (Ours)}
        & \textbf{0.26} (±0.02) & \textbf{1.40} (±0.08) & 
         \textbf{0.13} (±0.01) & \underline{30.08} (±0.05) & \underline{75.62} (±0.71) &  \underline{0.69} (±0.01) & \textbf{226.13} (±2.42) & \textbf{1.35} (±0.01)\\
        
        \bottomrule
    \end{tabular}}
\end{table*}
\begin{table*}[t]
    \caption{\textbf{Quantitative results on race attributes using Stable Diffusion v2.1.}~\cite{sd}
    Best and second-best results are indicated by \textbf{bold} and \underline{underline}.}
    \label{tab:main2}
    \resizebox{\linewidth}{!}{
    \begin{tabular}{c|ccc|ccccc}
        \toprule
        & \multicolumn{8}{c}{\textbf{Race}} \\
        \cmidrule(lr){2-9} 
        
        \textbf{Method}
        & \multicolumn{3}{c|}{\textbf{Fairness}}
        & \multicolumn{5}{c}{\textbf{Quality}} \\
        
        & Bias-O$\downarrow$ & Bias-Q$\downarrow$
        & FD$\downarrow$ & CLIP-T$\uparrow$ & CLIP-I$\uparrow$ & DINO$\uparrow$ & FID$\downarrow$ & IS$\uparrow$\\
        \midrule
        
        SD\cite{sd}
        & 0.63 (±0.01) & 2.06 (±0.35) & 0.21 (±0.01) & \underline{29.90} (±0.15) & - & - & 259.36 (±4.81) & 1.23 (±0.03)\\
        
        debias VL\cite{chuang2023debiasvl}
        & 0.49 (±0.03) & 1.91 (±0.92) & 0.14 (±0.01) & 28.15 (±0.26) & 67.42 (±0.96) & 0.46 (±0.02) & 242.78 (±4.21) & 1.33 (±0.03)\\
        
        UCE\cite{gandikota2024uce}
        & 0.50 (±0.03) & 1.95 (±0.37) & 0.16 (±0.01) & 29.44 (±0.12) & \textbf{80.46} (±1.13) & \underline{0.64} (±0.02) & 250.57 (±4.49) & 1.23 (±0.03)\\
        
        EntiGen\cite{entigen}
        & 0.55 (±0.03) & 3.07 (±0.39) & 0.14 (±0.01) & 28.12 (±0.12) & 65.34 (±1.02) & 0.45 (±0.02) & 253.53 (±3.83) & 1.23 (±0.03)\\
        
        LightFair\cite{lightfair}
        & \underline{0.40} (±0.03) & \underline{1.82} (±0.44) & 0.11 (±0.01) & \textbf{30.26} (±0.16) & \underline{77.47} (±1.05) & 0.53 (±0.03) & \underline{230.59} (±6.53) & \underline{1.35} (±0.01)\\
        
        \midrule
        \textbf{TES (Ours)}
        & \textbf{0.15} (±0.02) & \textbf{0.86} (±0.04) & \underline{0.09} (±0.01) & 29.49 (±0.07) & 69.87 (±0.35) & \textbf{0.65} (±0.01) & \textbf{228.52} (±1.08) & \textbf{1.36} (±0.01)\\
        
        \bottomrule
    \end{tabular}}
\end{table*}

\subsection{Experimental Setup}

\noindent{\textbf{Experimental Settings.}}
We conduct all experiments on Stable Diffusion v2.1~\cite{sd} using DDIM sampling with 40 inference steps and a classifier-free guidance (CFG) scale of 7.5.
Following standard evaluation settings in recent fairness studies on text-to-image diffusion models, we use the neutral prompt template \textit{``Photo portrait of a \{profession\}, a person''} and evaluate on six profession categories: \{artist, CEO, doctor, nurse, taxi driver, teacher\}.
We consider two demographic attributes: gender (\{male, female\}) and race (\{white, black, asian\}).
For each profession, we generate 100 images, resulting in 600 images per evaluation run, and repeat the experiment five times with different random seeds to account for stochastic variation in diffusion sampling.
All methods are evaluated under the same protocol, with baseline details provided in the supplementary material.
For fairness evaluation, we adopt Bias-O, Bias-Q, and FD.
Bias-O measures imbalance in the frequency of generated demographic attributes, while Bias-Q measures disparities in generation quality across attributes.
In addition, FD serves as a complementary fairness metric by measuring distributional discrepancy between generated attribute distributions.
For quality evaluation, we report CLIP-T, CLIP-I, DINO, FID, and IS.
Among them, CLIP-T measures text-image semantic alignment, CLIP-I measures consistency with the original model outputs, DINO captures feature-level similarity, FID evaluates image realism at the distribution level, and IS reflects image quality and diversity.
Demographic labels for fairness-related metrics are assigned using the FairFace classifier.
Samples without detected faces are excluded only from fairness-related metrics, and the same filtering rule is applied to all methods.

\subsection{Quantitative Results}

\noindent{\textbf{Overall Comparison.}}
Table~\ref{tab:main} and Table~\ref{tab:main2} report quantitative results on both gender and race debiasing using Stable Diffusion v2.1.
Overall, our method achieves the strongest fairness performance while preserving competitive, and in several cases superior, generation quality.
Across both demographic attributes, TES consistently ranks first or second on most fairness and quality metrics, demonstrating a favorable trade-off between demographic alignment and image quality.

\noindent{\textbf{Fairness Analysis.}}
Our method substantially improves fairness across all three fairness metrics: Bias-O, Bias-Q, and FD.
For gender debiasing, TES reduces Bias-O from 0.85 for the original Stable Diffusion baseline to 0.26, and lowers FD from 0.54 to 0.13.
For race debiasing, the improvements are even more pronounced: Bias-O decreases from 0.63 to 0.15, Bias-Q decreases from 2.06 to 0.86, and FD is reduced from 0.21 to 0.09.
These results indicate that our method not only reduces the dominance of majority demographic groups in generated samples, but also improves balance in the quality of generated images across attributes.

Compared with strong baselines such as LightFair, our method further improves fairness in both settings.
For example, under race debiasing, TES improves Bias-O from 0.40 to 0.15 and Bias-Q from 1.82 to 0.86 relative to LightFair, while also achieving a lower FD.
These improvements demonstrate that TES consistently reduces demographic imbalance beyond existing training-free baselines while maintaining competitive generation quality.

\noindent{\textbf{Quality Analysis.}}
Despite these strong fairness gains, our method maintains competitive generation quality.
In the gender setting, TES achieves a CLIP-T score of 30.08, which is close to the best-performing baseline and higher than the original Stable Diffusion model.
It also achieves the best FID score of 226.13, indicating that the generated image distribution remains highly realistic even after debiasing.
Similarly, in the race setting, TES obtains the best FID score of 228.52 and the highest IS score of 1.36, demonstrating that fairness improvement does not come at the cost of reduced visual quality or diversity.

The DINO results further support this observation.
TES achieves 0.69 for gender and 0.65 for race, outperforming or matching the strongest baselines in feature-level similarity.
Since DINO captures structural and perceptual consistency from image features rather than direct text-image matching, these results suggest that our method preserves the overall semantic and visual coherence of the generated samples while adjusting demographic attributes.
In other words, the debiasing effect is not achieved by simply distorting outputs toward target labels, but by steering the generation process in a visually consistent manner.
\begin{figure}[t]
    \centering
    \includegraphics[width=\linewidth]{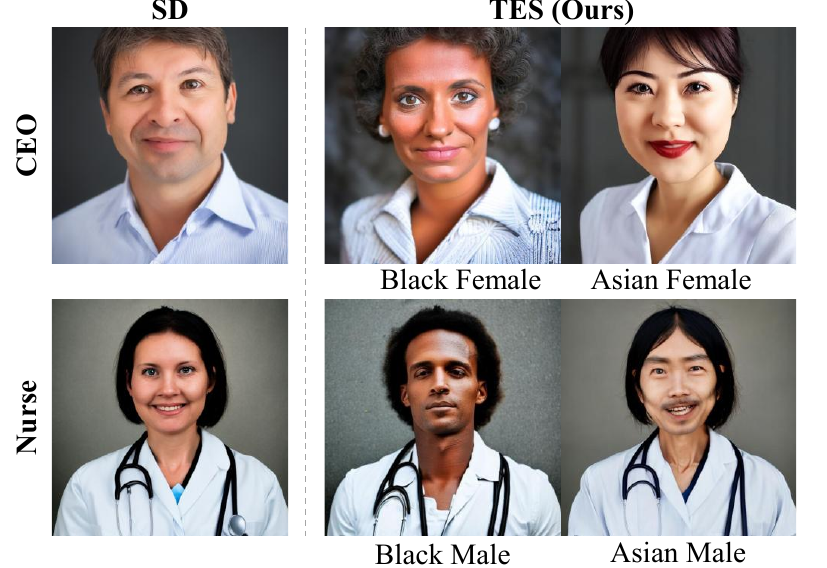}
    \caption{\textbf{Intersectional attribute control.}
    Our method enables simultaneous control of gender and race, generating diverse attribute combinations while preserving realism.}
    \label{fig:intersectional}
\end{figure}

\noindent{\textbf{Trade-off Between Fairness and Quality.}}
A notable trend in Table~\ref{tab:main} is that existing baselines often exhibit unstable trade-offs between fairness and quality.
For instance, some methods reduce Bias-O but still retain relatively high Bias-Q or FD, indicating that demographic frequencies may become more balanced while image quality remains uneven across groups.
Other methods preserve CLIP-I or DINO similarity well, but fail to sufficiently reduce demographic imbalance.
This inconsistency highlights a common challenge in bias mitigation for diffusion models: improving fairness along one axis often degrades semantic alignment, realism, or consistency along another.

In contrast, TES maintains a more balanced profile across all metrics.
Rather than achieving fairness through aggressive post-hoc editing or by relying on fixed control rules, our method continuously adjusts the conditional embedding in response to intermediate generation states.
This allows the model to better align demographic control with semantic preservation, resulting in lower fairness discrepancy together with strong CLIP-T, DINO, FID, and IS scores.


\begin{figure}[t]
    \centering
    \includegraphics[width=\linewidth]{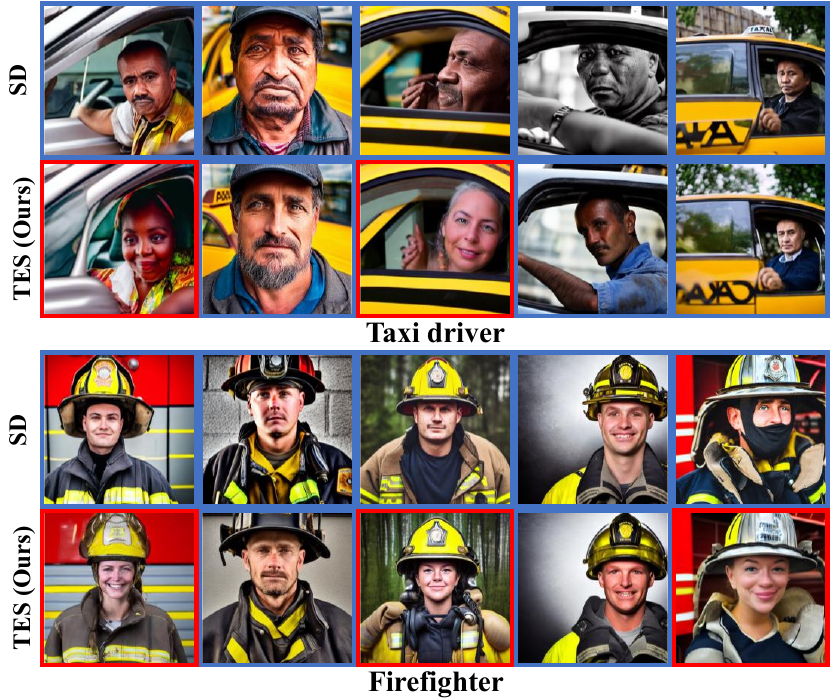}
    \caption{\textbf{Generalization to Stable Diffusion v1.5.}
    Our method consistently reduces bias while preserving semantic fidelity across different diffusion backbones.}
    \label{fig:sd15}
\end{figure}

\subsection{Qualitative Results}

Figure~\ref{fig:qualitative} presents qualitative comparisons across multiple professions.
The baseline tends to generate images concentrated on dominant demographic groups (e.g., male CEOs or female nurses).
In contrast, our method produces more balanced outputs while preserving identity-related features such as facial structure, pose, and overall scene composition.
As shown in Figure~\ref{fig:intersectional}, we further evaluate intersectional attribute control by jointly steering gender and race for highly biased professions, where our method generates a broader range of attribute combinations than the baseline while maintaining semantic consistency, demonstrating its effectiveness in multi-attribute bias mitigation without explicit supervision.
Finally, to assess generalization across diffusion backbones, we apply our method to Stable Diffusion v1.5 without modification. 
As shown in Figure~\ref{fig:sd15}, the baseline exhibits similar bias patterns as in Stable Diffusion v2.1, while our method consistently steers generated images toward the target attributes without degrading visual quality, indicating that the proposed approach generalizes effectively in a plug-and-play manner.

\subsection{Ablation Study}

\begin{table}[h]
    \caption{\textbf{Effect of different update strategies.} Initial update alone improves alignment but severely degrades fairness.}
    \label{tab:ablation_update}
    \resizebox{\linewidth}{!}{
    \begin{tabular}{l|cccccc}
        \toprule
        \textbf{Method} 
        & CLIP-T$\uparrow$ & Bias-O$\downarrow$ & Bias-Q$\downarrow$ & FD$\downarrow$ & CLIP-I$\uparrow$ & FID$\downarrow$ \\
        \midrule
        SD 
        & 29.86 & 0.82 & \textbf{1.06} & 0.41 & - & 246.61 \\
        Initial Only 
        & \textbf{30.17} & 0.81 & 5.76 & 0.40 & \textbf{87.97} & 236.38 \\
        Iterative Only 
        & 30.12 & \underline{0.29} & 1.53 & \underline{0.14} & \underline{77.44} & \underline{225.05} \\
        \textbf{Ours} 
        & \underline{30.14} & \textbf{0.25} & \underline{1.44} & \textbf{0.12} & 75.93 & \textbf{224.60} \\
        \bottomrule
    \end{tabular}}
\end{table}

\noindent{\textbf{Effect of Update Strategy.}} Table~\ref{tab:ablation_update} analyzes the role of each stage in the proposed two-stage optimization framework. 
Using only the initial update yields the highest CLIP-T and CLIP-I scores, indicating that early correction can effectively steer the global semantic trajectory. 
However, this strategy fails to achieve stable fairness control: Bias-Q rises sharply to 5.76, while Bias-O and FD remain close to the baseline. This suggests that a single early intervention is insufficient for stable debiasing.
In contrast, using only iterative refinement substantially improves fairness, reducing Bias-O from 0.82 to 0.29 and FD from 0.41 to 0.14. 
This confirms that repeated refinement during denoising is effective for correcting biased attribute emergence. 
The full two-stage strategy achieves the best overall trade-off by combining both effects: the early-stage update provides global directional correction, while iterative refinement preserves this correction as semantic details emerge. 
As a result, TES reduces Bias-O to 0.25 and FD to 0.12 while also achieving the best FID. 
These results show that reliable debiasing requires both coarse early-stage steering and progressive denoising-time refinement.
    
\begin{table}[H]
    \caption{\textbf{Effect of different loss components.} The full objective best balances alignment, fairness, and visual quality.}
    \label{tab:ablation_loss}
    \resizebox{\linewidth}{!}{
    \begin{tabular}{l|ccccccc}
        \toprule
        \textbf{Method} 
        & CLIP-T$\uparrow$ & Bias-O$\downarrow$ & Bias-Q$\downarrow$ & FD$\downarrow$ & CLIP-I$\uparrow$ & DINO$\uparrow$ & FID$\downarrow$ \\
        \midrule
        Attribute Loss Only 
        & 28.68 & 0.27 & 1.55 & 0.14 & 74.54 & 0.68 & \textbf{222.76} \\
        + L2 Loss 
        & 28.68 & 0.26 & 1.68 & 0.13 & 74.70 & 0.68 & 225.80 \\
        + Profession Loss 
        & \underline{30.13} & \textbf{0.25} & \textbf{1.22} & \underline{0.13} & \textbf{75.96} & \textbf{0.69} & 225.12 \\
        \textbf{Ours} 
        & \textbf{30.14} & \textbf{0.25} & \underline{1.44} & \textbf{0.12} & \underline{75.93} & \textbf{0.69} & \underline{224.60} \\
        \bottomrule
    \end{tabular}}
\end{table}

\noindent{\textbf{Effect of Loss Design.}}
Table~\ref{tab:ablation_loss} examines the contribution of each loss term in the optimization objective.
When only the gender objective is used, the model already improves fairness relative to the baseline, but the overall alignment remains weak, as reflected by the relatively low CLIP-T score of 28.68.
This indicates that optimizing only for the target demographic direction can correct attribute imbalance to some extent, but does not sufficiently preserve the intended profession semantics.
Adding L2 regularization slightly stabilizes the optimization and improves Bias-O and FD, but its effect remains limited.
Its role is mainly to prevent excessive drift of the optimized embedding rather than to directly improve fairness or alignment.
In contrast, the profession-preserving loss introduces a larger improvement: CLIP-T increases from 28.68 to 30.13, while Bias-O, Bias-Q, CLIP-I, and DINO also improve. 
This shows that preserving profession semantics is essential for robust debiasing. 
Overall, the full model achieves the most balanced result across demographic alignment, semantic preservation, and visual quality, demonstrating that effective inference-time debiasing requires jointly optimizing demographic control, semantic consistency, and embedding regularization.

\subsection{Discussion}
While the results demonstrate that TES effectively improves demographic balance under our evaluation protocol, TES is designed to steer generated samples toward a predefined demographic distribution rather than to provide a comprehensive fairness solution.
Additional implementation details, including inference-time analysis and experimental settings, are provided in the supplementary material.
We also include additional qualitative comparisons on Stable Diffusion v2.1, comprehensive evaluations on Stable Diffusion v1.5 against existing methods, and further ablation results.
Finally, extending TES to broader intersectional settings with multiple simultaneously varying attributes remains an important direction for future work.
\section{Conclusion}
In this paper, we proposed a training-free debiasing framework that operates at inference time by directly optimizing the conditional text embedding of diffusion models. 
Our method introduces a two-stage strategy that combines early-stage global alignment with iterative refinement during denoising, enabling stable and controllable attribute steering.
Extensive experiments demonstrate that our approach consistently reduces demographic bias across both gender and race while preserving image quality and semantic consistency.
In addition, qualitative and intersectional results show that the proposed method can control multiple demographic attributes without retraining or architectural modification.
Overall, our work highlights the effectiveness of text embedding optimization as a practical and scalable solution for fairness-aware generation in diffusion models.
\clearpage
{
    \small
    \bibliographystyle{ieeenat_fullname}
    \bibliography{main}

@String(CVPR= {IEEE Conf. Comput. Vis. Pattern Recog.})

@String(ICCV= {Int. Conf. Comput. Vis.})

@String(NIPS= {Adv. Neural Inform. Process. Syst.})

@String(ACMMM= {ACM Int. Conf. Multimedia})

@String(ICLR = {Int. Conf. Learn. Represent.})

@String(AAAI = {AAAI})

@String(CVPR  = {CVPR})

@String(ICCV  = {ICCV})

@String(NIPS  = {NeurIPS})

@String(ACMMM = {ACM MM})

@String(ICLR  = {ICLR})

@inproceedings{pmlr-v139-ramesh21a,
  author    = {Aditya Ramesh and Mikhail Pavlov and Gabriel Goh and Scott Gray and Chelsea Voss and Alec Radford and Mark Chen and Ilya Sutskever},
  title     = {Zero-Shot Text-to-Image Generation},
  booktitle = {ICML},
  year      = {2021}
}

@inproceedings{ddim,
  author = {Jiaming Song and Chenlin Meng and Stefano Ermon},
  title = {Denoising Diffusion Implicit Models},
  booktitle = ICLR,
  year = {2021}
}

@inproceedings{ddpm,
      title={Denoising Diffusion Probabilistic Models}, 
      author={Jonathan Ho and Ajay Jain and Pieter Abbeel},
      year={2020},
      booktitle = NIPS
}

@inproceedings{NEURIPS2022_ec795aea,
  author    = {Chitwan Saharia and William Chan and Saurabh Saxena and Lala Li and Jay Whang and Emily L. Denton and Kamyar Ghasemipour and Raphael Gontijo Lopes and Burcu Karagol Ayan and Tim Salimans and Jonathan Ho and David J. Fleet and Mohammad Norouzi},
  title     = {Photorealistic Text-to-Image Diffusion Models with Deep Language Understanding},
  booktitle = NIPS,
  year      = {2022}
}

@inproceedings{sd,
  author    = {Robin Rombach and Andreas Blattmann and Dominik Lorenz and Patrick Esser and Bj{\"o}rn Ommer},
  title     = {High-Resolution Image Synthesis with Latent Diffusion Models},
  booktitle = CVPR,
  year      = {2022}
}

@inproceedings{nichol2021glide,
  author        = {Alex Nichol and Prafulla Dhariwal and Aditya Ramesh and Pranav Shyam and Pamela Mishkin and Bob McGrew and Ilya Sutskever and Mark Chen},
  title         = {GLIDE: Towards Photorealistic Image Generation and Editing with Text-Guided Diffusion Models},
  year          = {2021},
  booktitle = {ICML}
}

@inproceedings{clip,
  author    = {Alec Radford and Jong Wook Kim and Chris Hallacy and Aditya Ramesh and Gabriel Goh and Sandhini Agarwal and Girish Sastry and Amanda Askell and Pamela Mishkin and Jack Clark and Gretchen Krueger and Ilya Sutskever},
  title     = {Learning Transferable Visual Models From Natural Language Supervision},
  booktitle = {ICML},
  year      = {2021}
}

@inproceedings{Cho_2023_ICCV,
  author    = {Jaemin Cho and Abhay Zala and Mohit Bansal},
  title     = {DALL-Eval: Probing the Reasoning Skills and Social Biases of Text-to-Image Generation Models},
  booktitle = ICCV,
  year      = {2023}
}

@inproceedings{wang-etal-2023-t2iat,
  author    = {Jialu Wang and Xinyue Liu and Zonglin Di and Yang Liu and Xin Wang},
  title     = {{T}2{IAT}: Measuring Valence and Stereotypical Biases in Text-to-Image Generation},
  booktitle = {Findings of the Association for Computational Linguistics (ACL)},
  year      = {2023}
}

@inproceedings{NEURIPS2023_b01153e7,
  author    = {Sasha Luccioni and Christopher Akiki and Margaret Mitchell and Yacine Jernite},
  title     = {Stable Bias: Evaluating Societal Representations in Diffusion Models},
  booktitle = NIPS,
  year      = {2023}
}

@inproceedings{Bianchi_2023,
  author     = {Federico Bianchi and Pratyusha Kalluri and Esin Durmus and Faisal Ladhak and Myra Cheng and Debora Nozza and Tatsunori Hashimoto and Dan Jurafsky and James Zou and Aylin Caliskan},
  title      = {Easily Accessible Text-to-Image Generation Amplifies Demographic Stereotypes at Large Scale},
  booktitle  = {2023 ACM Conference on Fairness, Accountability, and Transparency},
  year       = {2023},
  pages      = {1493--1504}
}

@article{wu2024stablediffusionexposedgender,
  author        = {Yankun Wu and Yuta Nakashima and Noa Garcia},
  title         = {Stable Diffusion Exposed: Gender Bias from Prompt to Image},
  year          = {2024},
  journal = {arXiv preprint arXiv:2312.03027},
}

@article{seshadri2023biasamplificationparadoxtexttoimage,
  author        = {Preethi Seshadri and Sameer Singh and Yanai Elazar},
  title         = {The Bias Amplification Paradox in Text-to-Image Generation},
  year          = {2023},
  journal = {arXiv preprint arXiv:2308.00755}
}

@article{culturalbias,
  author    = {Lukas Struppek and Dom Hintersdorf and Felix Friedrich and Manuel Brack and Patrick Schramowski and Kristian Kersting},
  title     = {Exploiting Cultural Biases via Homoglyphs in Text-to-Image Synthesis},
  journal   = {Journal of Artificial Intelligence Research},
  year      = {2023},
  pages     = {1017--1068}
}

@article{fairdiffusion,
  author        = {Felix Friedrich and Manuel Brack and Lukas Struppek and Dominik Hintersdorf and Patrick Schramowski and Sasha Luccioni and Kristian Kersting},
  title         = {Fair Diffusion: Instructing Text-to-Image Generation Models on Fairness},
  year          = {2023},
  journal       = {arXiv preprint arXiv:2302.10893}
}

@inproceedings{fairgen,
  author    = {Mintong Kang and Vinayshekhar Bannihatti Kumar and Shamik Roy and Abhishek Kumar and Sopan Khosla and Balakrishnan Murali Narayanaswamy and Rashmi Gangadharaiah},
  title     = {FairGen: Controlling Sensitive Attributes for Fair Generations in Diffusion Models via Adaptive Latent Guidance},
  booktitle = {Proceedings of the Conference on Empirical Methods in Natural Language Processing (EMNLP)},
  year      = {2025},
  pages     = {25336--25350}
}

@inproceedings{fu2025fairimagen,
  author    = {Zihao Fu and Ryan Brown and Shun Shao and Kai Rawal and Eoin D. Delaney and Chris Russell},
  title     = {FairImagen: Post-Processing for Bias Mitigation in Text-to-Image Models},
  booktitle = NIPS,
  year      = {2025}
}

@inproceedings{zhang2023itigen,
  author    = {Cheng Zhang and Xuanbai Chen and Siqi Chai and Henry Chen Wu and Dmitry Lagun and Thabo Beeler and Fernando De la Torre},
  title     = {{ITI-GEN}: Inclusive Text-to-Image Generation},
  booktitle = ICCV,
  year      = {2023}
}

@article{li2024fairmapping,
  author        = {Li, Jia and Hu, Lijie and Zhang, Jingfeng and Zheng, Tianhang and Zhang, Hua and Wang, Di},
  title         = {Fair Text-to-Image Diffusion via Fair Mapping},
  journal       = {arXiv preprint arXiv:2311.17695},
  year          = {2024},
  eprint        = {2311.17695},
  archivePrefix = {arXiv},
  primaryClass  = {cs.CV}
}

@inproceedings{shen2024finetuning,
  author    = {Xudong Shen and Chao Du and Tianyu Pang and Min Lin and Yongkang Wong and Mohan Kankanhalli},
  title     = {Finetuning Text-to-Image Diffusion Models for Fairness},
  booktitle = ICLR,
  year      = {2024}
}

@inproceedings{jiang2024laag,
  author    = {Yue Jiang and Yueming Lyu and Ziwen He and Bo Peng and Jing Dong},
  title     = {Mitigating Social Biases in Text-to-Image Diffusion Models via Linguistic-Aligned Attention Guidance},
  booktitle = ACMMM,
  year      = {2024}
}

@inproceedings{shrestha2024fairrag,
  author    = {Robik Shrestha and Yang Zou and Qiuyu Chen and Zhiheng Li and Yusheng Xie and Siqi Deng},
  title     = {FairRAG: Fair Human Generation via Fair Retrieval Augmentation},
  booktitle = CVPR,
  year      = {2024}
}

@inproceedings{parihar2024balancingact,
  author    = {Rishubh Parihar and Abhijnya Bhat and Abhipsa Basu and Saswat Mallick and Jogendra Nath Kundu and R. Venkatesh Babu},
  title     = {Balancing Act: Distribution-Guided Debiasing in Diffusion Models},
  booktitle = CVPR,
  year      = {2024}
}

@article{yesiltepe2024mist,
  author        = {Hidir Yesiltepe and Kiymet Akdemir and Pinar Yanardag},
  title         = {MIST: Mitigating Intersectional Bias with Disentangled Cross-Attention Editing in Text-to-Image Diffusion Models},
  year          = {2024},
  journal       = {arXiv preprint arXiv:2403.19738},
  eprint        = {2403.19738},
  archivePrefix = {arXiv},
  primaryClass  = {cs.CV},
}

@inproceedings{choi2024fairsm,
  author    = {Choi, Yujin and Park, Jinseong and Kim, Hoki and Lee, Jaewook and Park, Saerom},
  title     = {Fair Sampling in Diffusion Models through Switching Mechanism},
  booktitle = AAAI,
  year      = {2024},
}

@inproceedings{lightfair,
  author    = {Boyu Han and Qianqian Xu and Shilong Bao and Zhiyong Yang and Kangli Zi and Qingming Huang},
  title     = {LightFair: Towards an Efficient Alternative for Fair T2I Diffusion via Debiasing Pre-trained Text Encoders},
  booktitle = {Advances in Neural Information Processing Systems (NeurIPS)},
  year      = {2025}
}

@article{hou2026aitti,
  author        = {Xinyu Hou and Xiaoming Li and Chen Change Loy},
  title         = {AITTI: Learning Adaptive Inclusive Token for Text-to-Image Generation},
  year          = {2026},
  journal       = {arXiv preprint arXiv:2406.12805},
  eprint        = {2406.12805},
  archivePrefix = {arXiv},
  primaryClass  = {cs.CV},
}

@article{debiaspi,
      title={DebiasPI: Inference-time Debiasing by Prompt Iteration of a Text-to-Image Generative Model}, 
      author={Sarah Bonna and Yu-Cheng Huang and Ekaterina Novozhilova and Sejin Paik and Zhengyang Shan and Michelle Yilin Feng and Ge Gao and Yonish Tayal and Rushil Kulkarni and Jialin Yu and Nupur Divekar and Deepti Ghadiyaram and Derry Wijaya and Margrit Betke},
      year={2025},
      journal={arXiv preprint arXiv:2501.18642},
      eprint={2501.18642},
      archivePrefix={arXiv},
      primaryClass={cs.CV},
      url={https://arxiv.org/abs/2501.18642}, 
}

@article{moesd,
      title={MoESD: Mixture of Experts Stable Diffusion to Mitigate Gender Bias}, 
      author={Guorun Wang and Lucia Specia},
      year={2024},
      journal={arXiv preprint arXiv:2407.11002},
      eprint={2407.11002},
      archivePrefix={arXiv},
      primaryClass={cs.CL},
      url={https://arxiv.org/abs/2407.11002}, 
}

@InProceedings{Liu_2024_CVPR,
    author    = {Liu, Zhixuan and Schaldenbrand, Peter and Okogwu, Beverley-Claire and Peng, Wenxuan and Yun, Youngsik and Hundt, Andrew and Kim, Jihie and Oh, Jean},
    title     = {SCoFT: Self-Contrastive Fine-Tuning for Equitable Image Generation},
    booktitle = CVPR,
    year      = {2024}
}

@inproceedings{park2025efa,
  author    = {Jeonghoon Park and Juyoung Lee and Chaeyeon Chung and Jaeseong Lee and Jaegul Choo and Jindong Gu},
  title     = {Fair Generation without Unfair Distortions: Debiasing Text-to-Image Generation with Entanglement-Free Attention},
  booktitle = {Proceedings of the IEEE/CVF International Conference on Computer Vision (ICCV)},
  year      = {2025}
}

@inproceedings{vardhana2026fully,
  author    = {Korada Sri Vardhana and Shrikrishna Lolla and Soma Biswas},
  title     = {Fully Unsupervised Self-debiasing of Text-to-Image Diffusion Models},
  booktitle = {WACV},
  year      = {2026}
}

@inproceedings{gandikota2024uce,
  author    = {Rohit Gandikota and Hadas Orgad and Yonatan Belinkov and Joanna Materzy{\'n}ska and David Bau},
  title     = {Unified Concept Editing in Diffusion Models},
  booktitle = {Proceedings of the IEEE/CVF Winter Conference on Applications of Computer Vision (WACV)},
  year      = {2024},
  pages     = {5111--5120}
}

@inproceedings{orgad2023time,
  author    = {Hadas Orgad and Bahjat Kawar and Yonatan Belinkov},
  title     = {Editing Implicit Assumptions in Text-to-Image Diffusion Models},
  booktitle = {Proceedings of the IEEE/CVF International Conference on Computer Vision (ICCV)},
  year      = {2023},
  pages     = {7053--7061}
}

@article{saedebias,
      title={Model-Agnostic Gender Bias Control for Text-to-Image Generation via Sparse Autoencoder}, 
      author={Chao Wu and Zhenyi Wang and Kangxian Xie and Naresh Kumar Devulapally and Vishnu Suresh Lokhande and Mingchen Gao},
      year={2025},
      journal={arXiv preprint arXiv:2507.20973},
      eprint={2507.20973},
      archivePrefix={arXiv},
      primaryClass={cs.LG},
      url={https://arxiv.org/abs/2507.20973}, 
}

@article{chuang2023debiasvl,
    author = {Ching-Yao Chuang and Varun Jampani and Yuanzhen Li and Antonio Torralba and Stefanie Jegelka},
    title = {Debiasing vision-language models via biased prompts},
    year = {2023},
    journal = {arXiv preprint arXiv:2302.00070}
}

@inproceedings{entigen,
  author    = {Hritik Bansal and Da Yin and Masoud Monajatipoor and Kai-Wei Chang},
  title     = {How Well Can Text-to-Image Generative Models Understand Ethical Natural Language Interventions?},
  booktitle = {Proceedings of the Conference on Empirical Methods in Natural Language Processing (EMNLP)},
  year      = {2022},
  pages     = {1358--1370}
}

@inproceedings{fid,
  author    = {Martin Heusel and Hubert Ramsauer and Thomas Unterthiner and Bernhard Nessler and Sepp Hochreiter},
  title     = {GANs Trained by a Two Time-Scale Update Rule Converge to a Local Nash Equilibrium},
  booktitle = NIPS,
  year      = {2017}
}

@inproceedings{is,
  author    = {Tim Salimans and Ian Goodfellow and Wojciech Zaremba and Vicki Cheung and Alec Radford and Xi Chen},
  title     = {Improved Techniques for Training GANs},
  booktitle = NIPS,
  year      = {2016}
}

@article{dino,
  author  = {Maxime Oquab and Timoth{\'e}e Darcet and Th{\'e}o Moutakanni and Huy Vo and Marc Szafraniec and Vasil Khalidov and Pierre Fernandez and Daniel Haziza and Francisco Massa and Alaaeldin El-Nouby and others},
  title   = {DINOv2: Learning Robust Visual Features Without Supervision},
  journal = {Transactions on Machine Learning Research (TMLR)},
  year    = {2024},
  pages   = {1--31}
}

@inproceedings{fd,
  author    = {Kristy Choi and Aditya Grover and Trisha Singh and Rui Shu and Stefano Ermon},
  title     = {Fair Generative Modeling via Weak Supervision},
  booktitle = {ICML},
  year      = {2020}
}

@inproceedings{textual_inversion,
  author    = {Rinon Gal and Yuval Alaluf and Yuval Atzmon and Or Patashnik and Amit Haim Bermano and Gal Chechik and Daniel Cohen-Or},
  title     = {An Image is Worth One Word: Personalizing Text-to-Image Generation Using Textual Inversion},
  booktitle = ICLR,
  year      = {2023}
}

@inproceedings{saner,
  author    = {Yusuke Hirota and Min-Hung Chen and Chien-Yi Wang and Yuta Nakashima and Yu-Chiang Frank Wang and Ryo Hachiuma},
  title     = {SANER: Annotation-Free Societal Attribute Neutralizer for Debiasing CLIP},
  booktitle = ICLR,
  year      = {2025}
}

@inproceedings{dear,
  author    = {Ashish Seth and Mayur Hemani and Chirag Agarwal},
  title     = {DEAR: Debiasing Vision-Language Models with Additive Residuals},
  booktitle = CVPR,
  year      = {2023}
}

@inproceedings{finetune_fair,
  author    = {Xudong Shen and Chao Du and Tianyu Pang and Min Lin and Yongkang Wong and Mohan Kankanhalli},
  title     = {Finetuning Text-to-Image Diffusion Models for Fairness},
  booktitle = ICLR,
  year      = {2024}
}

@article{perera2023analyzing,
    author = {Malsha V. Perera and Vishal M. Patel},
    title = {Analyzing Bias in Diffusion-Based Face Generation Models},
    journal = {arXiv preprint arXiv:2305.06402},
    year = {2023}
}

@article{jain2020imperfect,
    author = {Niharika Jain and Alberto Olmo and Sailik Sengupta and Lydia Manikonda and Subbarao Kambhampati},
    title = {Imperfect {ImaGANation}: Implications of {GANs} Exacerbating Biases on Facial Data Augmentation and {Snapchat} Selfie Lenses},
    journal = {arXiv preprint arXiv:2001.09528},
    year = {2020}
}

@inproceedings{vanbreugel2021decaf,
    author = {Boris van Breugel and Trent Kyono and Jeroen Berrevoets and Mihaela van der Schaar},
    title = {{DECAF}: Generating Fair Synthetic Data Using Causally-Aware Generative Networks},
    booktitle = NIPS,
    year = {2021}
}

@inproceedings{teo2023fairtl,
  author    = {Christopher T. H. Teo and Milad Abdollahzadeh and Ngai-Man Cheung},
  title     = {Fair Generative Models via Transfer Learning},
  booktitle = AAAI,
  year      = {2023}
}
}

\clearpage
\makeatletter
\let\Hy@backout\@gobble
\makeatother

\clearpage
\setcounter{page}{1}
\setcounter{section}{0}

\maketitlesupplementary

\section{Implementation Details}
This section provides additional implementation details that are omitted from the main paper for brevity. Unless otherwise specified, all experiments follow the settings described in Sec.~5.1 of the main paper.



\subsection{Attribute and Semantic Alignment}

At every active optimization timestep, the reconstructed clean image $\hat{x}_0$ is decoded into the image space and evaluated using CLIP~\cite{clip}. Both image and text features are $\ell_2$-normalized before cosine similarity is computed.

For gender debiasing, the candidate attribute descriptions consist of a target and an opposite attribute (e.g., ``male'' and ``female''). For race debiasing, the candidate set is extended to three classes (``Asian'', ``Black'', and ``White''), where the optimization objective encourages the generated image toward the assigned target class while suppressing the remaining candidates. The final objective combines the attribute alignment loss, profession-preserving loss, and embedding regularization described in Sec.~4.2 of the main paper.

\paragraph{Target Attribute Assignment.}
Target attributes are assigned automatically using a lightweight preview stage. 
For each random seed, a low-cost preview image is generated using a reduced number of diffusion steps. 
As shown in Table~\ref{tab:runtime}, adding the preview stage increases inference time only slightly, from 36.61s to 37.29s per image, while leaving peak GPU memory unchanged.
CLIP similarity scores are then computed for all candidate demographic descriptions. 
Samples are ranked according to these scores and assigned to demographic targets using a balanced allocation strategy. For gender, the ranked samples are divided equally into male and female targets. 
For race, class-wise quotas are used to obtain a balanced distribution across the three demographic groups. 
This procedure is performed once before the main optimization and does not require intermediate clean-image reconstruction.

\subsection{Optimization Details}

As described in Sec.~5.1 of the main paper, we adopt DDIM sampling with 40 denoising steps for all experiments. Embedding optimization is activated only within the selected timestep window corresponding to 10\%--50\% of the denoising trajectory (steps 4--19 for 40 sampling steps). In addition, TES performs an initial embedding update before the main optimization stage by rolling out the latent trajectory to 5\% of the denoising process and applying a small number of gradient updates with a reduced learning rate. Gradient clipping is applied during both the initial update and denoising-time optimization to improve numerical stability.

\subsection{Computational Overhead}

Table~\ref{tab:runtime} reports the computational overhead of TES measured on a single NVIDIA RTX 4090 GPU at $512\times512$ image resolution and 40 DDIM sampling steps. Wall-clock inference time is averaged after one warm-up run, and peak GPU memory is measured using \texttt{torch.cuda.max\_memory\_allocated()}. The preview stage adds only a small overhead, increasing the generation-only runtime from 36.61s to 37.29s per image. The additional cost mainly comes from repeated CLIP evaluation and gradient-based text-embedding updates, while the diffusion backbone remains frozen.

\begin{table}[t]
\centering
\renewcommand{\thetable}{S\arabic{table}}
\setcounter{table}{0}
\caption{\textbf{Inference-time computational overhead.}}
\label{tab:runtime}
\resizebox{\linewidth}{!}{
\begin{tabular}{lccc}
\toprule
Method & Time / Image (s)$\downarrow$ & Peak GPU Memory (GB)$\downarrow$ & Relative Time$\downarrow$ \\
\midrule
Stable Diffusion & 3.19 & 3.59 & 1.00$\times$ \\
TES (w/o Preview) & 36.61 & 9.87 & 11.48$\times$ \\
TES (w/ Preview) & 37.29 & 9.87 & 11.69$\times$ \\
\bottomrule
\end{tabular}
}
\end{table}

\section{Additional Results}
\subsection{Quantitative Results}

\begin{figure}[t]
    \centering
    \renewcommand{\thefigure}{S\arabic{figure}}
    \setcounter{figure}{0}
    \includegraphics[width=\linewidth]{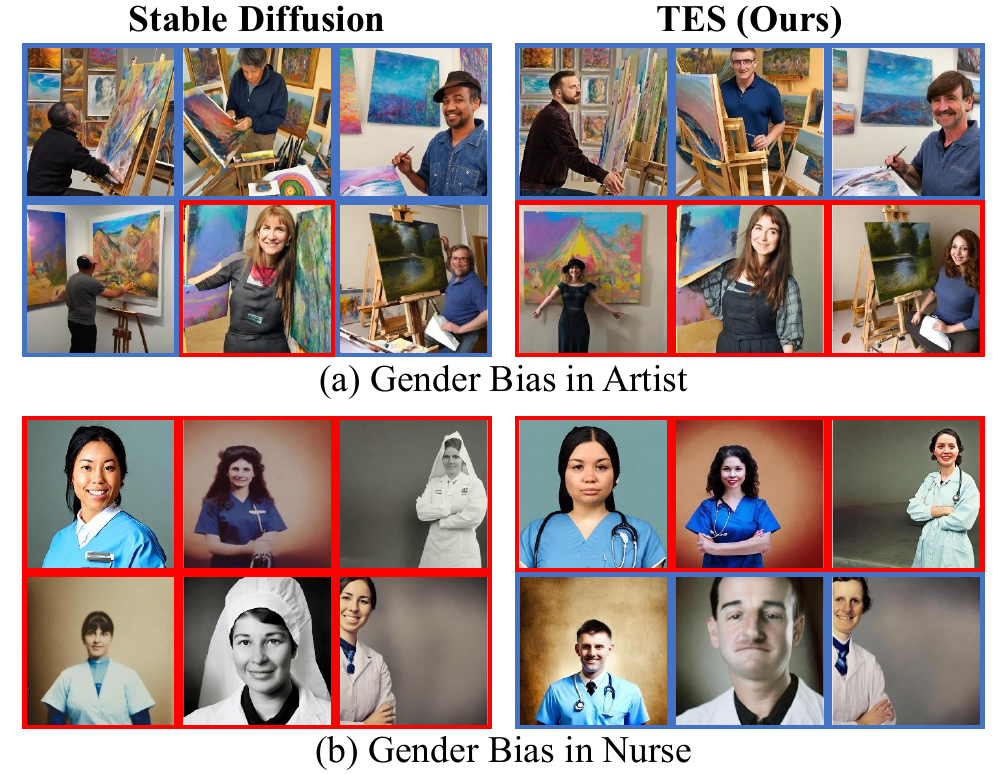}
    \caption{\textbf{Qualitative results on unseen prompt templates for gender attributes using Stable Diffusion v2.1.} Compared with the baseline Stable Diffusion model, TES produces more balanced gender representations while preserving profession semantics. Colored borders indicate gender categories: blue denotes male and red denotes female.}
    \label{fig:new_prompt_gender}
\end{figure}
\begin{figure}[t]
    \centering
    \renewcommand{\thefigure}{S\arabic{figure}}
    \setcounter{figure}{1}
    \includegraphics[width=\linewidth]{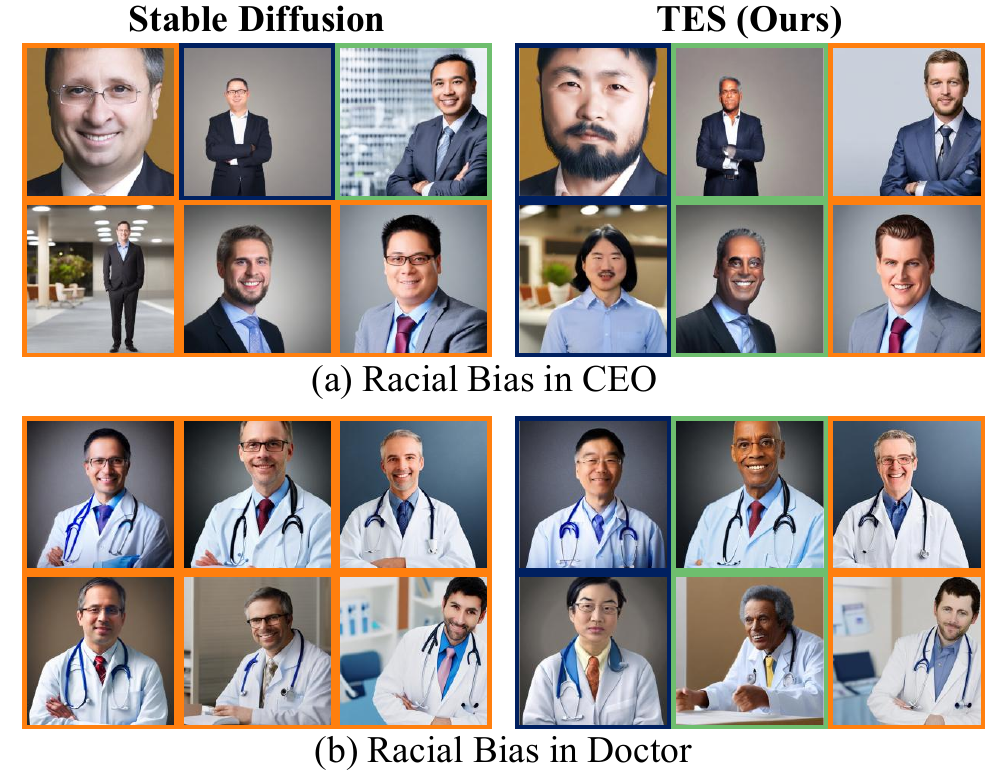}
    \caption{\textbf{Qualitative results on unseen prompt templates for race attributes using Stable Diffusion v2.1.} Compared with the baseline Stable Diffusion model, TES produces more balanced racial representations while preserving profession semantics. Colored borders indicate race categories: navy denotes Asian, green denotes Black, and orange denotes White.}
    \label{fig:new_prompt_race}
\end{figure}

\paragraph{Generalization to Other Backbones.}
To assess whether our method is specific to a particular diffusion model, we additionally evaluate it on Stable Diffusion v1.5 under the same experimental protocol.
As shown in Table~\ref{tab:sd15_gender} and Table~\ref{tab:sd15_race}, our approach consistently improves fairness metrics while maintaining competitive generation quality, following the same trends observed in Stable Diffusion v2.1.
These results indicate that our method is not tied to a specific backbone, but can be applied to different diffusion models in a plug-and-play manner.
Importantly, the relative improvements over baselines remain consistent across both models.

\paragraph{Generalization to Unseen Prompt Templates.}
We further evaluate prompt robustness by replacing the main prompt template with \textit{``a photo of a \{profession\}''}.
As shown in Tables~\ref{tab:prompt_gender} and~\ref{tab:prompt_race}, TES consistently improves all fairness metrics over the Stable Diffusion baseline under this alternative prompt formulation.
At the same time, the quality metrics remain within a comparable range, indicating that the method does not achieve debiasing by simply sacrificing generation quality.
These results suggest that the proposed inference-time optimization generalizes beyond the prompt template used in the main experiments.

\subsection{Qualitative Results}
\paragraph{Qualitative Results across Diffusion Backbones.}
We present qualitative comparisons between the baseline Stable Diffusion model and our method on both Stable Diffusion v2.1 and v1.5.
As shown in Figures~\ref{fig:sd21_qual} and~\ref{fig:sd15_qual}, the baseline models exhibit clear demographic bias patterns, including gender imbalance and racial concentration across profession prompts such as \textit{artist}, \textit{doctor}, \textit{teacher}, \textit{nurse}, and \textit{CEO}.
In contrast, our method produces more diverse and balanced demographic representations while preserving the semantic consistency of the given profession.
These results qualitatively demonstrate that the proposed inference-time embedding optimization mitigates both gender and racial biases without degrading visual quality, and generalizes across diffusion backbones without architectural modification.

\subsection{Prompt and Profession Generalization}
\textbf{Generalization to Unseen Prompt Templates.}
The main experiments use the prompt template \textit{``Photo portrait of a \{profession\}, a person''}.
To evaluate whether our method is dependent on a specific prompt formulation, we additionally replace the template with \textit{``a photo of a \{profession\}''}.
As shown in Figures~\ref{fig:new_prompt_gender} and~\ref{fig:new_prompt_race}, the baseline Stable Diffusion model continues to exhibit pronounced gender and racial biases under the alternative prompt formulation.
In contrast, TES consistently produces more balanced gender and racial representations while preserving the intended profession semantics.
Furthermore, Figure~\ref{fig:new_prof} shows qualitative results on an additional set of profession prompts that are not used in the main experiments, such as \textit{fashion designer}, \textit{hairdresser}, \textit{lawyer}, \textit{technician}, \textit{librarian}, and \textit{pilot}.
TES also yields more balanced gender representations on these additional professions, indicating that the proposed inference-time embedding optimization generalizes not only to unseen prompt templates but also to profession categories beyond those used in the main experiments.

\begin{table}[t]
    \centering
    \renewcommand{\thetable}{S\arabic{table}}
    \setcounter{table}{1}
    \caption{\textbf{Quantitative results on gender attributes under an alternative prompt template using Stable Diffusion v2.1.}}
    \label{tab:prompt_gender}
    \resizebox{\linewidth}{!}{
    \begin{tabular}{c|ccc|ccccc}
    \toprule
    \textbf{Method}
    & \multicolumn{3}{c|}{\textbf{Fairness}}
    & \multicolumn{5}{c}{\textbf{Quality}} \\
    
    & Bias-O$\downarrow$ & Bias-Q$\downarrow$ & FD$\downarrow$
    & CLIP-T$\uparrow$ & CLIP-I$\uparrow$ & DINO$\uparrow$ & FID$\downarrow$ & IS$\uparrow$ \\
    \midrule
    
    SD~\cite{sd}
    & 0.73 & 1.90 & 0.45
    & \textbf{29.31} & - & - & \textbf{275.13} & 1.26 \\
     
    \textbf{Ours}
    & \textbf{0.19} & \textbf{1.65} & \textbf{0.10}
    & 28.88 & 75.47 & 0.66 & 303.99 & \textbf{1.43} \\
    
    \bottomrule
    \end{tabular}}
\end{table}

\begin{table}[t]
    \centering
    \renewcommand{\thetable}{S\arabic{table}}
    \setcounter{table}{2}
    \caption{\textbf{Quantitative results on race attributes under an alternative prompt template using Stable Diffusion v2.1.}}
    \label{tab:prompt_race}
    \resizebox{\linewidth}{!}{
    \begin{tabular}{c|ccc|ccccc}
    \toprule
    \textbf{Method}
    & \multicolumn{3}{c|}{\textbf{Fairness}}
    & \multicolumn{5}{c}{\textbf{Quality}} \\
    
    & Bias-O$\downarrow$ & Bias-Q$\downarrow$ & FD$\downarrow$
    & CLIP-T$\uparrow$ & CLIP-I$\uparrow$ & DINO$\uparrow$ & FID$\downarrow$ & IS$\uparrow$ \\
    \midrule
    
    SD~\cite{sd}
    & 0.54 & 1.60 & 0.17
    & \textbf{29.31} & - & - & \textbf{275.13} & 1.26 \\
     
    \textbf{Ours}
    & \textbf{0.14} & \textbf{1.14} & \textbf{0.08}
    & 28.64 & 70.45 & 0.59 & 307.35  & \textbf{1.40} \\
    
    \bottomrule
    \end{tabular}}
\end{table}

\begin{table*}[p]
    \centering
    \renewcommand{\thetable}{S\arabic{table}}
    \setcounter{table}{3}
    \caption{\textbf{Quantitative results on gender attributes using Stable Diffusion v1.5.} Our method consistently improves fairness while maintaining competitive generation quality, following trends similar to those observed in Stable Diffusion v2.1.}
    \label{tab:sd15_gender}
    \resizebox{\linewidth}{!}{
    \begin{tabular}{c|ccc|ccccc}
    \toprule
    \textbf{Method}
    & \multicolumn{3}{c|}{\textbf{Fairness}}
    & \multicolumn{5}{c}{\textbf{Quality}} \\
    
    & Bias-O$\downarrow$ & Bias-Q$\downarrow$ & FD$\downarrow$
    & CLIP-T$\uparrow$ & CLIP-I$\uparrow$ & DINO$\uparrow$ & FID$\downarrow$ & IS$\uparrow$ \\
    \midrule
    
    SD~\cite{sd} & 0.73 (±0.05) & 1.90 (±0.67) & 0.45 (±0.03)
    & 29.31 (±0.06) & - & - & 275.13 (±6.75) & 1.26 (±0.03) \\
    
    FairD~\cite{fairdiffusion} & 0.79 (±0.04) & 3.25 (±1.15) & 0.45 (±0.02)
    & 28.79 (±0.11) & 75.91 (±0.56) & 0.53 (±0.02) & 269.62 (±4.42) & \underline{1.30} (±0.03) \\
    
    UCE~\cite{gandikota2024uce} & 0.78 (±0.07) & 1.79 (±0.46) & 0.48 (±0.04)
    & 28.91 (±0.13) & \textbf{82.72} (±0.81) & \textbf{0.70} (±0.02) & 273.95 (±5.53) & 1.26 (±0.03) \\
    
    FinetuneFD~\cite{finetune_fair} & 0.38 (±0.07) & 2.31 (±0.35) & 0.22 (±0.04)
    & \underline{29.34} (±0.13) & 76.17 (±0.68) & 0.57 (±0.01) & 278.21 (±7.53) & 1.24 (±0.02) \\
    
    FairMapping~\cite{li2024fairmapping} & 0.46 (±0.05) & 2.16 (±0.72) & 0.30 (±0.03)
    & 29.30 (±0.16) & 76.00 (±0.66) & 0.53 (±0.02) & 278.81 (±5.84) & 1.26 (±0.02) \\
    
    BalancingAct~\cite{parihar2024balancingact} & 0.41 (±0.05) & 1.70 (±0.55) & 0.24 (±0.03)
    & 29.30 (±0.11) & 77.37 (±0.64) & 0.55 (±0.02) & 272.08 (±5.16) & 1.28 (±0.02) \\
    
    TI~\cite{textual_inversion} & 0.56 (±0.06) & 1.88 (±0.37) & 0.33 (±0.04)
    & 28.76 (±0.10) & 75.43 (±0.54) & 0.54 (±0.02) & 278.92 (±6.22) & 1.27 (±0.02) \\
    
    AITTI~\cite{hou2026aitti} & 0.41 (±0.06) & 1.34 (±0.44) & 0.25 (±0.02)
    & 29.03 (±0.09) & 77.25 (±0.44) & 0.56 (±0.02) & 267.23 (±5.14) & 1.29 (±0.02) \\
    
    TIME~\cite{orgad2023time} & 0.65 (±0.04) & 1.76 (±0.35) & 0.40 (±0.02)
    & 28.45 (±0.12) & 73.71 (±0.69) & 0.52 (±0.02) & 279.17 (±4.43) & 1.25 (±0.02) \\
    
    MIST~\cite{yesiltepe2024mist} & 0.39 (±0.05) & 1.35 (±0.43) & 0.22 (±0.03)
    & 29.10 (±0.13) & 76.67 (±0.35) & 0.55 (±0.01) & 254.33 (±4.76) & 1.27 (±0.02) \\
    
    FairSM~\cite{choi2024fairsm}& 0.65 (±0.04) & 1.83 (±0.21) & 0.43 (±0.02)
    & 27.98 (±0.11) & 74.23 (±0.59) & 0.52 (±0.02) & 265.60 (±5.04) & 1.26 (±0.03) \\
    
    SANER~\cite{saner} & 0.52 (±0.02) & 1.65 (±0.34) & 0.35 (±0.02)
    & 28.13 (±0.08) & 75.28 (±0.77) & 0.52 (±0.02) & 275.34 (±5.40) & 1.25 (±0.04) \\
    
    DEAR~\cite{dear} & 0.73 (±0.05) & 1.90 (±0.67) & 0.45 (±0.03)
    & 29.31 (±0.06) & - & - & 275.13 (±6.75) & 1.26 (±0.03) \\
    
    EntiGen~\cite{entigen} & 0.46 (±0.05) & 2.63 (±0.88) & 0.27 (±0.03)
    & 28.57 (±0.14) & 71.89 (±0.68) & 0.50 (±0.01) & 263.76 (±5.51) & 1.29 (±0.04) \\
    
    ITI-GEN~\cite{zhang2023itigen} & 0.39 (±0.06) & \underline{1.27} (±0.82) & 0.18 (±0.04)
    & 28.36 (±0.12) & 68.82 (±0.59) & 0.45 (±0.02) & \underline{246.55} (±5.97) & 1.29 (±0.02) \\
    
    LightFair~\cite{lightfair} & \underline{0.30} (±0.08) & \textbf{0.99} (±0.55)
    & \underline{0.17} (±0.04) & \textbf{30.57} (±0.16) & \underline{80.09} (±0.76) & \underline{0.63} (±0.02)
    & \textbf{233.53} (±5.50) & \underline{1.30} (±0.03) \\
    
    \midrule
    \textbf{Ours} & \textbf{0.09 (±0.01)} & 1.88 (±0.10) & \textbf{0.05 (±0.01)} 
    & 28.68 (±0.10) & 73.93 (±0.35) & 0.64 (±0.01) & 253.04 (±2.20) & \textbf{1.36 (±0.01)} \\
    
    \bottomrule
    \end{tabular}}
\end{table*}

\begin{table*}[p]
    \centering
    \renewcommand{\thetable}{S\arabic{table}}
    \setcounter{table}{4}
    \caption{\textbf{Quantitative results on race attributes using Stable Diffusion v1.5.} Our method generalizes to multi-class attribute control and maintains balanced performance across demographic groups.}
    \label{tab:sd15_race}
    \resizebox{\linewidth}{!}{
    \begin{tabular}{c|ccc|ccccc}
    \toprule
    \textbf{Method}
    & \multicolumn{3}{c|}{\textbf{Fairness}}
    & \multicolumn{5}{c}{\textbf{Quality}} \\
    
    & Bias-O$\downarrow$ & Bias-Q$\downarrow$ & FD$\downarrow$
    & CLIP-T$\uparrow$ & CLIP-I$\uparrow$ & DINO$\uparrow$ & FID$\downarrow$ & IS$\uparrow$ \\
    \midrule
    
    SD~\cite{sd}
    & 0.54 (±0.02) & 1.60 (±0.67) & 0.17 (±0.01)
    & 29.31 (±0.06) & - & - & 275.13 (±6.75) & 1.26 (±0.03) \\
    
    FairD~\cite{fairdiffusion}
    & 0.50 (±0.02) & 1.50 (±0.38) & 0.15 (±0.01)
    & 28.95 (±0.10) & 74.33 (±0.68) & 0.53 (±0.02) & 262.72 (±4.84) & 1.28 (±0.03) \\
    
    UCE~\cite{gandikota2024uce}
    & 0.44 (±0.03) & 1.40 (±0.24) & 0.13 (±0.01)
    & 29.13 (±0.14) & \textbf{90.15} (±0.70) & \textbf{0.83} (±0.02) & 281.16 (±5.18) & 1.26 (±0.02) \\
    
    FinetuneFD~\cite{finetune_fair}
    & 0.20 (±0.03) & 1.41 (±0.23) & \underline{0.07} (±0.01)
    & 29.02 (±0.15) & 74.57 (±0.53) & 0.54 (±0.01) & 270.09 (±5.99) & 1.26 (±0.02) \\
    
    FairMapping~\cite{li2024fairmapping}
    & 0.34 (±0.02) & 1.75 (±0.47) & 0.10 (±0.01)
    & 29.29 (±0.15) & 76.54 (±0.71) & 0.54 (±0.02) & 280.95 (±5.02) & 1.26 (±0.03) \\
    
    BalancingAct~\cite{parihar2024balancingact}
    & 0.34 (±0.02) & 1.13 (±0.36) & \underline{0.07} (±0.01)
    & \underline{29.34} (±0.11) & 77.44 (±0.72) & 0.55 (±0.02) & 271.91 (±5.35) & 1.29 (±0.03) \\
    
    TI~\cite{textual_inversion}
    & 0.47 (±0.03) & 1.45 (±0.27) & 0.14 (±0.01)
    & 28.67 (±0.17) & 67.96 (±0.84) & 0.43 (±0.04) & 275.20 (±5.07) & 1.25 (±0.03) \\
    
    AITTI~\cite{hou2026aitti}
    & 0.25 (±0.04) & 1.20 (±0.31) & 0.08 (±0.02)
    & 29.03 (±0.11) & \underline{85.43} (±0.47) & \underline{0.79} (±0.01) & 271.13 (±5.24) & 1.29 (±0.01) \\
    
    TIME~\cite{orgad2023time}
    & 0.39 (±0.04) & 1.51 (±0.34) & 0.12 (±0.01)
    & 27.97 (±0.15) & 76.53 (±0.68) & 0.54 (±0.02) & 275.72 (±6.84) & 1.26 (±0.02) \\
    
    MIST~\cite{yesiltepe2024mist}
    & 0.26 (±0.03) & 1.19 (±0.24) & 0.08 (±0.01)
    & 29.08 (±0.09) & 83.25 (±0.75) & 0.76 (±0.02) & 265.83 (±6.78) & 1.28 (±0.02) \\
    
    FairSM~\cite{choi2024fairsm}
    & 0.42 (±0.03) & 1.61 (±0.37) & 0.13 (±0.01)
    & 28.68 (±0.06) & 72.83 (±0.60) & 0.51 (±0.03) & 271.58 (±5.83) & 1.27 (±0.03) \\
    
    SANER~\cite{saner}
    & 0.45 (±0.03) & 1.41 (±0.33) & 0.13 (±0.02)
    & 28.50 (±0.13) & 73.64 (±0.51) & 0.50 (±0.02) & 273.21 (±6.42) & 1.24 (±0.02) \\
    
    DEAR~\cite{dear}
    & 0.54 (±0.02) & 1.60 (±0.67) & 0.17 (±0.01)
    & 29.31 (±0.06) & - & - & 275.13 (±6.75) & 1.26 (±0.03) \\
    
    EntiGen~\cite{entigen}
    & 0.37 (±0.04) & 2.88 (±0.63) & 0.09 (±0.01)
    & 27.97 (±0.14) & 69.56 (±0.74) & 0.47 (±0.01) & 265.94 (±4.25) & 1.31 (±0.04) \\
    
    ITI-GEN~\cite{zhang2023itigen}
    & 0.31 (±0.04) & 1.62 (±0.37) & 0.10 (±0.01)
    & 28.13 (±0.16) & 66.97 (±0.57) & 0.42 (±0.01) & 269.84 (±6.61) & \underline{1.33} (±0.03) \\
    
    LightFair~\cite{lightfair}
    & \underline{0.18} (±0.04) & \underline{1.06} (±0.43)
    & \textbf{0.06 (±0.01)}
    & \textbf{31.34} (±0.20) & \underline{86.31} (±0.70)
    & \underline{0.81} (±0.02) & \underline{259.96} (±7.75) & \underline{1.33} (±0.03) \\
    
    \midrule
    \textbf{Ours}
    & \textbf{0.14 (±0.02)} & \textbf{1.03 (±0.11)} & 0.08 (±0.01)
    & 28.53 (±0.06) & 68.95 (±0.36) & 0.64 (±0.01) & \textbf{253.62 (±1.84)}  & \textbf{1.37 (±0.01)} \\
    
    \bottomrule
    \end{tabular}}
\end{table*}

\begin{figure*}[p]
    \centering
    \renewcommand{\thefigure}{S\arabic{figure}}
    \setcounter{figure}{2}
    \includegraphics[width=\linewidth]{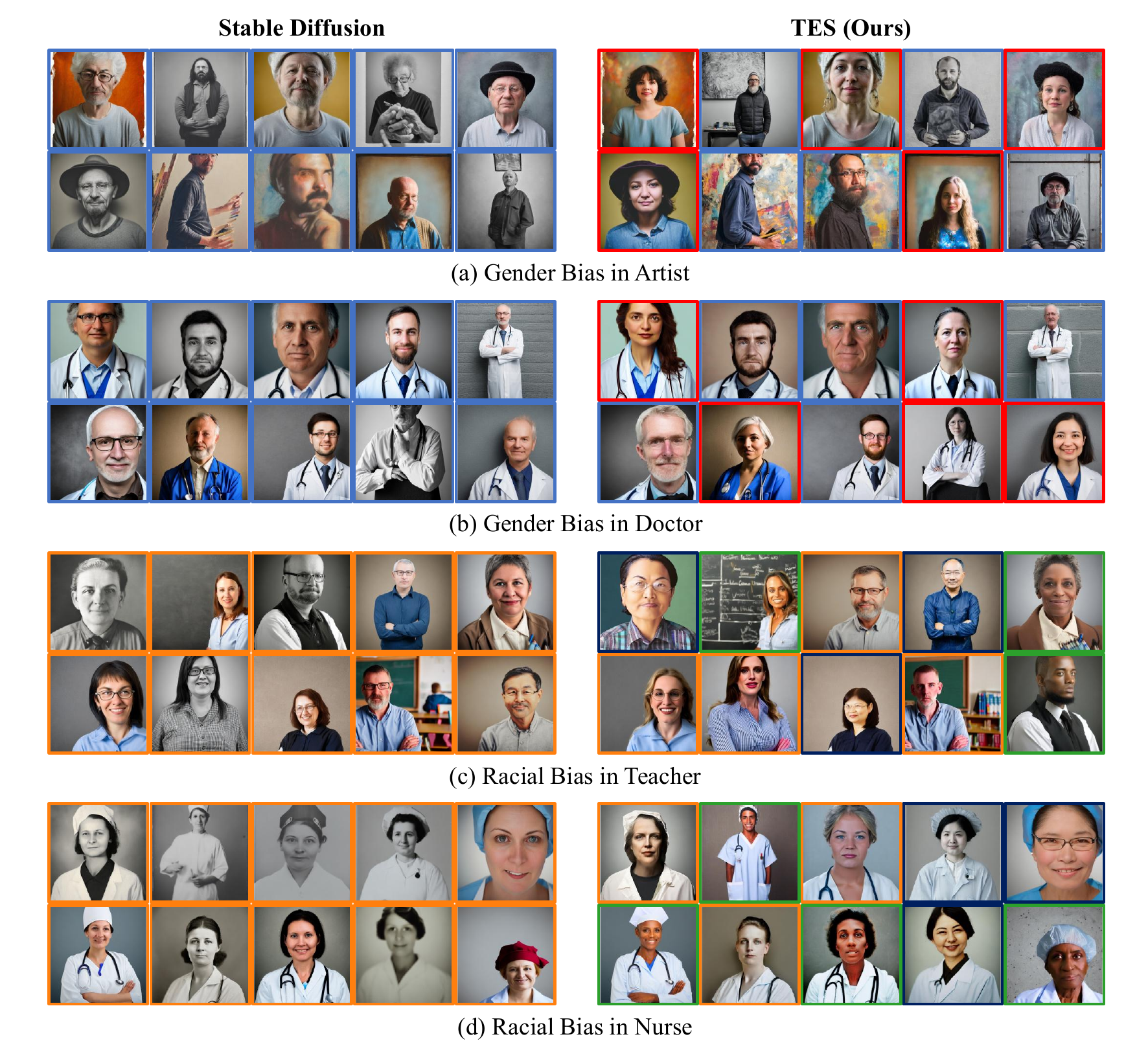}
    \caption{\textbf{Qualitative results on Stable Diffusion v2.1.} Compared with the baseline Stable Diffusion model, TES produces more balanced demographic representations across both gender and race while preserving profession semantics. Colored borders indicate attribute categories: for gender, blue denotes male and red denotes female; for race, navy denotes Asian, green denotes Black, and orange denotes White.}
    \label{fig:sd21_qual}
\end{figure*}

\begin{figure*}[p]
    \centering
    \renewcommand{\thefigure}{S\arabic{figure}}
    \setcounter{figure}{3}
    \includegraphics[width=\linewidth]{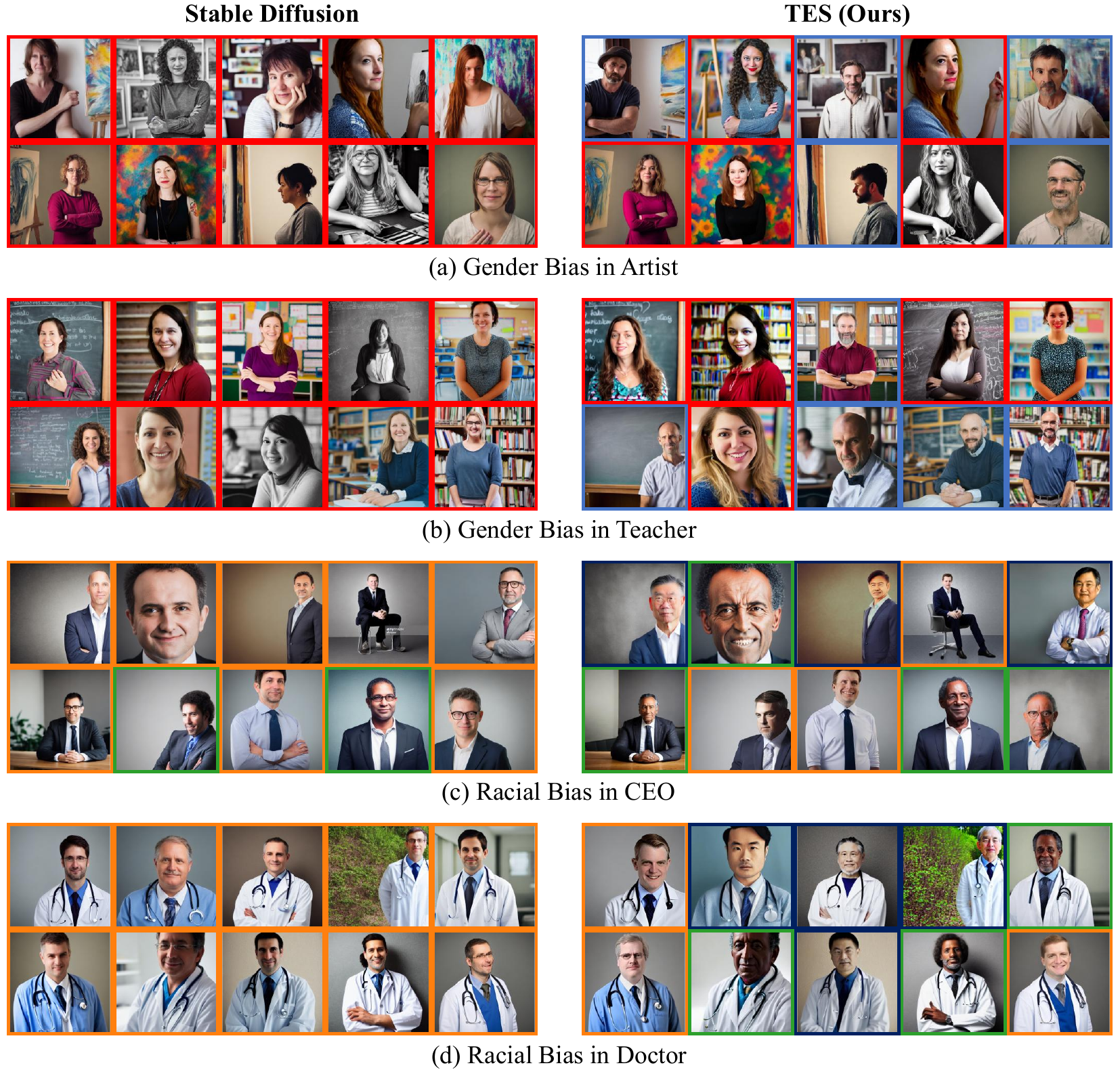}
    \caption{\textbf{Qualitative results on Stable Diffusion v1.5.} Compared with the baseline Stable Diffusion model, TES produces more balanced demographic representations across both gender and race while preserving profession semantics. Colored borders indicate attribute categories: for gender, blue denotes male and red denotes female; for race, navy denotes Asian, green denotes Black, and orange denotes White.}
    \label{fig:sd15_qual}
\end{figure*}

\begin{figure*}[p]
    \centering
    \renewcommand{\thefigure}{S\arabic{figure}}
    \setcounter{figure}{4}
    \includegraphics[width=\linewidth]{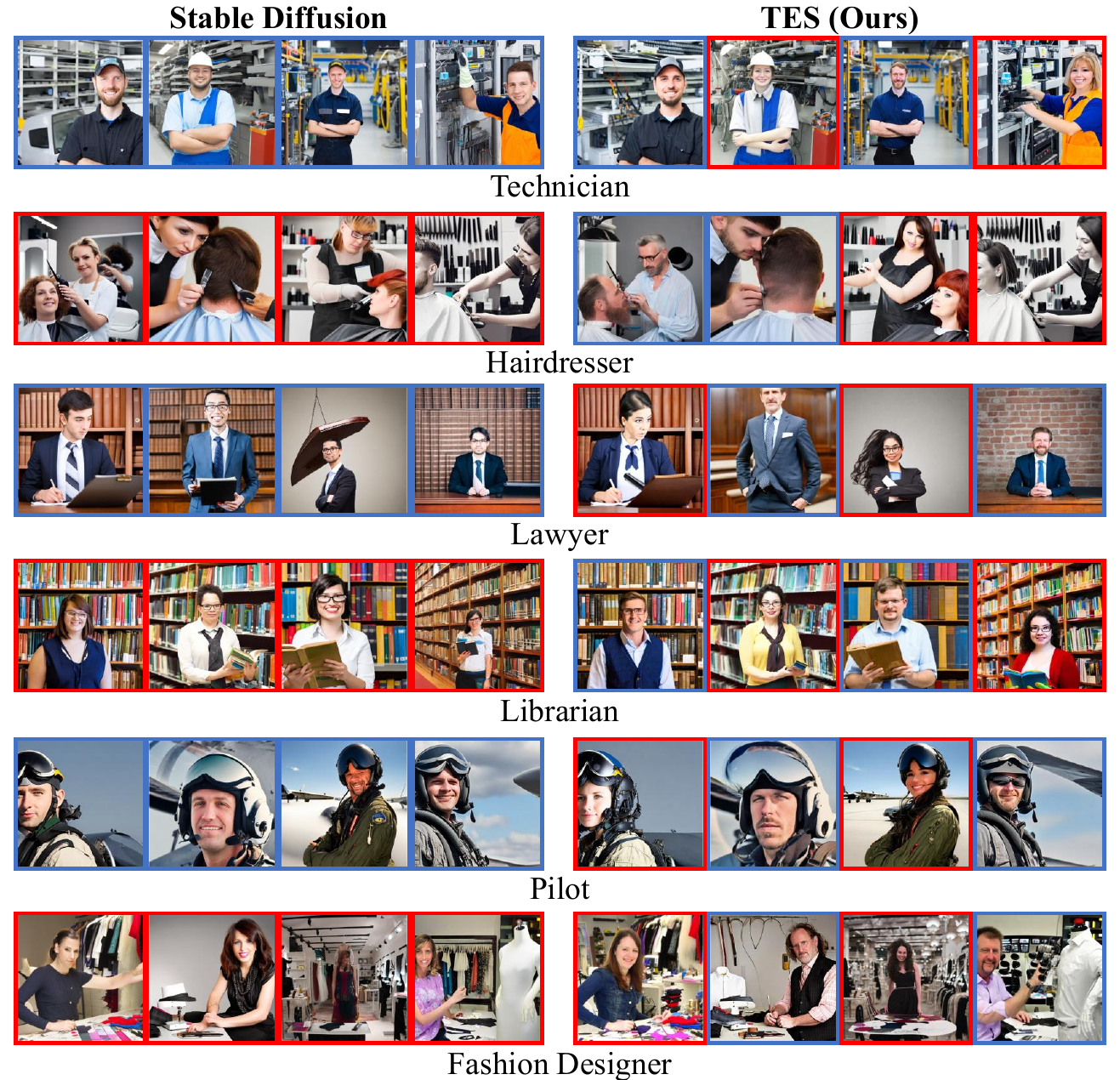}
    \caption{\textbf{Qualitative results on unseen professions using Stable Diffusion v2.1.} TES generalizes to profession categories not used in the main experiments, producing more balanced gender representations while preserving profession semantics. Colored borders indicate gender categories: blue denotes male and red denotes female.}
    \label{fig:new_prof}
\end{figure*}

\section{Fairness Metrics Details}

Following \textbf{LightFair}~\cite{lightfair}, we adopt the same evaluation protocol and report both \textit{fairness} and \textit{generation quality} metrics.
In our experiments, we consider six professions: \textit{doctor, CEO, taxi driver, nurse, artist,} and \textit{teacher}. For each profession, we generate 100 images, resulting in a total of 600 generated images for evaluation. All metrics are computed over this set.
The fairness evaluation is conducted based on predicted attribute labels (e.g., gender and race), while quality is measured using semantic alignment, perceptual similarity, and distributional statistics.

\subsection{Fairness Metrics}

We follow the LightFair protocol for fairness evaluation, where generated images are grouped according to predicted attributes, and fairness is measured in terms of both \textit{distribution balance} and \textit{consistency across groups}.

\begin{itemize}
    \item \textbf{Bias-O}~\cite{lightfair} evaluates whether unspecified attributes are generated with balanced frequency. Generated images are first assigned attribute labels using pretrained classifiers, and the frequency of each attribute is computed per prompt. Bias-O reflects the imbalance among these frequencies. Lower values indicate a more uniform distribution of attributes.

    \item \textbf{Bias-Q}~\cite{lightfair} evaluates whether generation quality is consistent across different attribute groups. After grouping images by predicted attributes, we compute a quality score for each group and measure discrepancies across groups. In our setup, the group-level quality is derived from CLIP-based text-image alignment (CLIP-T). Lower values indicate more consistent quality across attributes.

    \item \textbf{FD (Feature Distance)}~\cite{fd} measures the deviation between generated images and their corresponding baseline images in a feature space. Each generated image is paired with the baseline image produced using the same prompt and random seed, and the feature distance is computed and averaged across samples. Lower FD indicates better preservation of baseline characteristics.
\end{itemize}

\subsection{Quality Metrics}

To evaluate generation quality, we adopt the metrics used in LightFair and provide additional feature-based analysis.

\begin{itemize}
    \item \textbf{CLIP-T}~\cite{clip} measures text-image semantic alignment. For each generated image, we compute the CLIP score with respect to its corresponding prompt instantiated from a fixed template (\textit{``Photo portrait of a \{profession\}, a person''}). The scores are averaged per profession and then across all samples. Higher CLIP-T indicates better alignment between generated images and textual descriptions.

    \item \textbf{CLIP-I}~\cite{clip} measures image-level consistency with the original diffusion output. Each generated image is paired with its baseline counterpart generated using the same prompt and random seed. We extract normalized CLIP image features and compute cosine similarity. Higher CLIP-I indicates stronger similarity to the original output.

    \item \textbf{DINO}~\cite{dino} evaluates feature-level similarity using self-supervised visual representations. We extract global features from a pretrained DINOv2 model and compute cosine similarity between generated and baseline images. Higher DINO indicates better preservation of visual structure.

    \item \textbf{FID}~\cite{fid} measures the distributional similarity between generated images and real images. We compute FID with an Inception-v3 backbone. As the reference distribution, we use 10,000 images from the FairFace dataset, and compare them with the generated samples for each profession, followed by averaging across all professions. Lower FID indicates higher realism.

    \item \textbf{IS}~\cite{is} evaluates both image quality and diversity. Generated images are passed through a pretrained Inception network, and the score is computed using a split-based protocol. We report both mean and standard deviation across samples. Higher IS indicates better quality and diversity.
\end{itemize}

\end{document}